\documentclass{article} 
\usepackage{iclr2027_conference,times}

\usepackage{amsmath,amsfonts,bm}

\def\eqref#1{equation~\ref{#1}}

\def\1{\bm{1}}

\DeclareMathAlphabet{\mathsfit}{\encodingdefault}{\sfdefault}{m}{sl}
\SetMathAlphabet{\mathsfit}{bold}{\encodingdefault}{\sfdefault}{bx}{n}

\usepackage{hyperref}
\usepackage{url}
\usepackage{booktabs}
\usepackage{multirow}
\usepackage{graphicx} 

\title{Same molecule, Different Knowledge: Cross-Identifier Binding Failures in LLMs For chemistry}

\author{Khouloud Saadi\thanks{Corresponding author} , Di Wang \\
Department of Computer Science \\
King Abdullah University of Science and Technology\\
KAUST, Thuwal, Saudi Arabia \\
\texttt{\{khouloud.saadi,di.wang\}@kaust.edu.sa} \\
}

\iclrfinalcopy 
\begin{document}

\maketitle
\begin{abstract}
Chemical AI workflows routinely represent the same molecule through multiple
co-referring identifiers, including trivial names, IUPAC names, and SMILES,
implicitly assuming that these representations provide interchangeable access
to an LLM's knowledge. To test this
assumption, we first introduce a controlled benchmark and show that it does not hold. Across foundation models
spanning 7B to 72B parameters, even chemistry-specialized models,
co-referring molecular identifiers consistently expose different subsets of
the model's knowledge. 
We then conduct a series of controlled behavioral, representational, and
causal analyses to characterize this failure. We show that even when a model can identify a
molecule from its SMILES and retrieve a fact through the corresponding name,
the same fact can remain inaccessible directly from SMILES. We term this phenomenon \emph{cross-identifier binding failure}.   
Further analysis of internal representations, conflicting identifiers, and
parameter editing shows that neither improved chemical representation
competence nor modification of individual facts is sufficient to establish
reliable access across identifiers. Together, these results indicate that the
core failure lies in the binding between an identifier and knowledge already
present in the model. Motivated by this diagnosis, we introduce Binding
Self-Distillation, a simple mitigation that uses lexical identity during
training to transfer the model's existing knowledge access to the structural
route. This enables factual retrieval directly from SMILES at inference time,
without requiring a name lookup or training the underlying model from scratch.
\end{abstract}

\section{Introduction}

Large language models (LLMs) are increasingly used as general-purpose
interfaces to knowledge and reasoning~\citep{brown2020language,ouyang2022training,kojima2022large}.
Beyond natural-language generation~\citep{wei2022chain}, they support
applications in programming~\citep{chen2021evaluating},
medicine~\citep{mcduff2025towards,johri2025evaluation}, and scientific
discovery~\citep{singhal2023large,zhang2024comprehensive}. In chemistry,
LLMs are used for chemical question answering~\citep{zhang2024chemllm,zhao2025developing},
molecular property prediction~\citep{jablonka2024leveraging}, and molecule
design~\citep{bhattacharya2024large}. As these models become components of
chemical workflows, an important question is whether the knowledge they
contain can be accessed reliably under the different representations used to
describe the same molecule.

\begin{figure*}[t]
    \centering
    \includegraphics[width=0.9\textwidth]{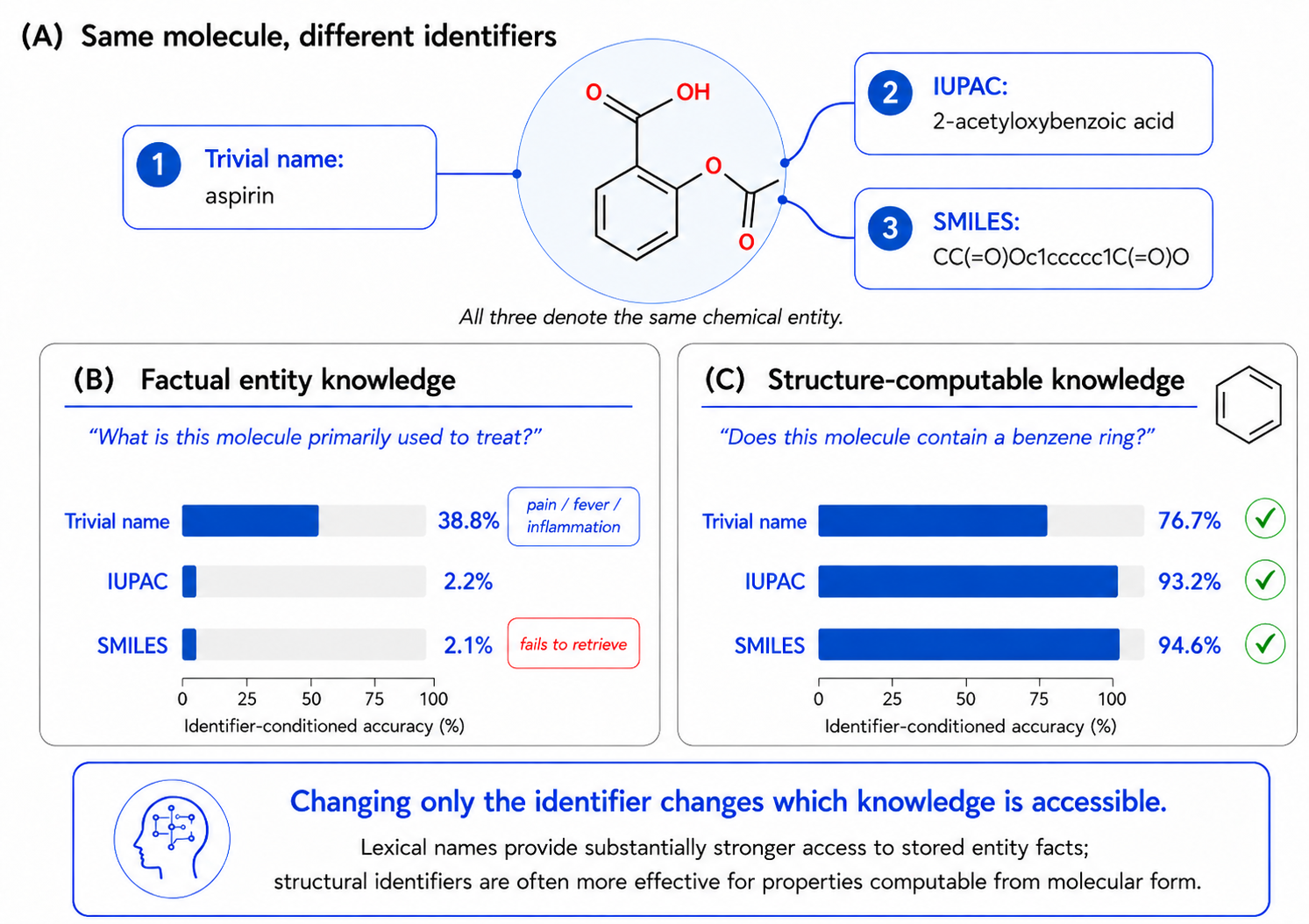}
    \caption{
    \textbf{Equivalent molecular identifiers expose different knowledge.}
    \textbf{(A)} A trivial name, IUPAC name, and SMILES string denote the same
    molecule (aspirin).
    \textbf{(B--C)} Identifier-conditioned accuracy over the 896 molecules in
    our benchmark, holding the molecule and queried property fixed while
    changing only the identifier. Lexical names provide substantially stronger
    access to entity-associated factual knowledge
    (\textbf{B}, therapeutic indication), whereas structural identifiers
    provide stronger access to structure-computable knowledge
    (\textbf{C}, benzene-ring presence). Representative results are shown for
    ChemDFM-v2.0-14B~\citep{zhao2025developing}.
    }
    \label{fig:intro}
    \vspace{-2 em}
\end{figure*}

A single molecule can be referred to through multiple co-referring
identifiers~\citep{kim2023pubchem}. For example, as shown in
Figure~\ref{fig:intro}(A), aspirin may appear as the trivial name
\textit{aspirin}, the systematic IUPAC name
\textit{2-acetyloxybenzoic acid}~\citep{favre2013nomenclature}, or the SMILES
string \texttt{CC(=O)Oc1ccccc1C(=O)O}~\citep{gomez2018automatic}.
These representations are routinely mixed across chemical workflows:
biomedical resources commonly use lexical names~\citep{hastings2016chebi},
whereas reaction datasets and molecular generative pipelines frequently use
structural representations such as SMILES~\citep{gomez2018automatic,schwaller2019molecular}.
Consequently, the same LLM may be queried about the same molecule through
different identifiers. This practice relies on an implicit assumption:
\emph{co-referring identifiers should be interchangeable routes to the
model's knowledge}. More generally, equivalence at the level of the
underlying scientific entity is implicitly expected to translate into
equivalence as an interface to the model's memory.

In this work, we first show that this assumption does not hold. We construct
\textbf{ChemIDBench}, a controlled benchmark for evaluating knowledge access
across co-referring molecular identifiers, covering approved drugs with matched
trivial names, IUPAC names, and SMILES,
and evaluate seven chemistry-specialized and general-purpose LLMs.
 Holding the underlying molecule
and queried property fixed while changing only its identifier reveals a
systematic double dissociation: as shown in Figure~\ref{fig:intro}(B,C), lexical names
provide substantially stronger access to entity-associated
factual knowledge, whereas
structure-descriptive identifiers provide stronger access to properties
computable from molecular form. Identifier choice therefore does more than
change the difficulty of a query; it changes \emph{which knowledge is
reachable}.
More surprisingly, this asymmetry persists even when the model recognizes the
molecule. We isolate cases in which a model can generate the correct trivial
name from SMILES and retrieve a fact through that name, yet still fails to
retrieve the same fact directly from SMILES. We call this phenomenon
\emph{cross-identifier binding failure}. It separates three notions that are
often implicitly conflated: recognizing the entity, possessing factual
knowledge about it, and binding a particular identifier to that knowledge.

We next ask what prevents knowledge from transferring across identifiers and
whether the missing access can be recovered. Supplying the
lexical name restores much of the inaccessible factual knowledge,
showing that the knowledge is present but not reached through the structural
route. More importantly, activation patching provides causal evidence for this
account: transplanting lexical-route states into a structural query restores
inaccessible factual retrieval under SMILES alone. Layer-wise analysis further shows that chemistry-specific
training can improve alignment between structural forms without establishing
comparable name-structure binding.
This weak binding is also a reliability problem: when conflicting identifiers
are presented together, different access routes can control different answers,
producing inconsistencies that models often fail to detect.
Supplying the lexical identity is an effective inference-time remedy, but it
requires an identifier lookup on every query. Standard knowledge editing does
not reliably propagate factual access across co-referring identifiers,
suggesting that modifying the factual association alone is insufficient.
Motivated by this diagnosis, we introduce Binding Self-Distillation, which
targets the missing access relation by transferring the model's own
name-conditioned access behavior to the SMILES route. It substantially improves
factual retrieval directly from SMILES across seven models, without requiring
the lexical name at inference time or training a model from scratch. A detailed discussion of related work is provided in Appendix~\ref{RL}. In summary, our contributions are:

\begin{itemize}
    \item We introduce \textbf{ChemIDBench}, a controlled benchmark for
    evaluating knowledge access across co-referring molecular identifiers,
    covering matched trivial names, IUPAC names, and SMILES and evaluating
    seven LLMs spanning 7B--72B parameters.

    \item We characterize how molecular identifier choice
    affects access to different types of knowledge, contrasting
    entity-associated factual queries with structure-computable queries and
    testing whether successful molecular recognition is sufficient for
    factual retrieval.

    \item Through lexical-bridge controls, conflicting-identifier experiments,
    representation analysis, and activation patching, we show that entity
    recognition, factual knowledge, and identifier-conditioned access are
    separable, providing behavioral and causal evidence for
    \emph{cross-identifier binding failure}.

    \item We show that standard knowledge editing does not reliably propagate
    factual access across co-referring identifiers, revealing that modifying
    the factual association alone is insufficient. Motivated by this diagnosis,
    we introduce Binding Self-Distillation, which directly targets the missing
    access relation and substantially improves retrieval from SMILES without
    requiring the lexical name at inference time.
\end{itemize}
\vspace{-1 em}

\section{Benchmarking Cross-Identifier Knowledge Access}
\vspace{-1 em}
\subsection{Benchmark Design}
\label{sec:benchmark_design}
\vspace{-0.5 em}
Our benchmark \textbf{ChemIDBench} isolates the effect of molecular identifier choice while holding
the underlying molecule and queried property fixed. Each molecule is represented by three co-referring molecular identifiers: a trivial name, an IUPAC name, and a canonical SMILES string. We select these
three because they are widely used across chemical and biomedical workflows
~\citep{hastings2016chebi,gomez2018automatic,schwaller2019molecular} while
providing complementary interfaces to the same molecular entity: trivial names
serve as familiar lexical references, IUPAC names provide systematic chemical
nomenclature~\citep{favre2013nomenclature}, and SMILES provides a compact
structure-based representation~\citep{gomez2018automatic}.
We begin with 3,025 molecules with approved
drug–indication records in ChEMBL~\citep{mendez2019chembl,hunter2025drug},
using the ChEMBL preferred name as the trivial-name identifier. We then apply a
strict entity-cleaning pipeline, removing salts, multi-fragment molecules,
duplicate names and structures, invalid structures, and extreme molecule
sizes, leaving 1,553 unambiguous molecular entities. We further retain only
molecules with scorable therapeutic indications, yielding 897 candidates, and
resolve their IUPAC names through PubChem~\citep{kim2023pubchem}. This produces
896 molecules with matched trivial-name, IUPAC-name, and SMILES representations.
Full filtering criteria, indication normalization, and identifier-resolution
checks are provided in Appendix~\ref{app:benchmark_details}.

For each task, we query the same molecule independently through its trivial
name, IUPAC name, and SMILES. The benchmark spans two complementary types of
queries: (1) \emph{Entity-associated queries} test access to information tied to
the molecular entity, including molecule identification and therapeutic
indication; (2) \emph{Structure-computable queries} test properties that can  be derived from molecular structure, including molecular formula,
benzene-ring presence, nitrogen-atom count, ring count, and heavy-atom count,
with ground truth computed using RDKit~\citep{rdkit}. Models use greedy
decoding; entity-associated outputs are scored against normalized accepted
answers, while structure-computable properties are evaluated against
RDKit-derived ground truth. Exact task definitions, prompts, ground-truth
construction, decoding settings, and scoring procedures are provided in
Appendix~\ref{app:evaluation_protocol1}.

We evaluate seven LLMs spanning 7B--72B parameters, including both
chemistry-specialized and general-purpose models. To isolate the effect
of chemistry-specific training on cross-identifier behavior, we compare
ChemDFM-v2.0-14B and ChemDFM-v1.5-8B~\citep{zhao2025developing}
with Qwen2.5-14B-Instruct~\citep{yang2024qwen25} and
Meta-Llama-3-8B-Instruct~\citep{grattafiori2024llama}, respectively.
We further include Qwen2.5-7B-Instruct~\citep{yang2024qwen25},
Llama-3.1-8B-Instruct~\citep{grattafiori2024llama}, and
Qwen2-72B-Instruct~\citep{yang2024qwen2}.
\vspace{-1 em}
\subsection{Equivalent Identifiers Expose Different Knowledge}
\label{sec:identifier_access}
\vspace{-0.5 em}
If co-referring identifiers were interchangeable interfaces to model
knowledge, changing only how a molecule is represented should have a negligible 
effect on the model's answer. Table~\ref{tab:identifier_access} shows
the opposite: changing the identifier can nearly erase access to
entity-associated knowledge while improving access to structure-computable
properties.
For ChemDFM-v2.0-14B, therapeutic-indication accuracy collapses from
$38.8\%$ under the trivial name to $2.1\%$ under SMILES. Yet on the
same molecules, the direction reverses for benzene-ring detection:
accuracy rises from $76.7\%$ to $94.6\%$. This reversal holds across all
seven models, spanning 7B to 72B parameters and including both
chemistry-specialized and general-purpose LLMs. Across these models, trivial
names are strongest for therapeutic indication, whereas SMILES outperforms
the trivial name on benzene detection by $13.3$--$28.3$ percentage points. Nitrogen counting shows the same broad
reversal, though the better structural representation, IUPAC or SMILES, varies across models.
Thus, structure-descriptive identifiers are not simply harder inputs: they can
be better interfaces when the requested information is recoverable from
molecular form.

\begin{table*}[t]
\centering
\small
\caption{
Accuracy (\%) under trivial name (N), IUPAC name (I), and SMILES (S).
\textit{Indication} denotes therapeutic-indication retrieval;
\textit{Benzene} denotes benzene-ring detection; and
\textit{Nitrogens} denotes nitrogen-atom counting.
These three tasks provide the primary evidence for the cross-identifier
dissociation; results for all seven benchmark tasks are reported in
Appendix~\ref{app:identifier_access_details}.
Best performance within each model and task is shown in bold.
}
\label{tab:identifier_access}
\resizebox{\textwidth}{!}{
\begin{tabular}{lccc|ccc|ccc}
\toprule
& \multicolumn{3}{c|}{\textbf{Indication}}
& \multicolumn{3}{c|}{\textbf{Benzene}}
& \multicolumn{3}{c}{\textbf{Nitrogens}} \\
\textbf{Model} & N & I & S & N & I & S & N & I & S \\
\midrule

ChemDFM-v2.0-14B
& \textbf{38.8} & 2.2 & 2.1
& 76.7 & 93.2 & \textbf{94.6}
& 41.6 & \textbf{70.3} & 59.6 \\

ChemDFM-v1.5-8B
& \textbf{22.5} & 0.1 & 3.7
& 70.5 & 81.1 & \textbf{84.4}
& 5.5 & 8.7 & \textbf{21.2} \\

Qwen2.5-14B-Instruct
& \textbf{37.8} & 5.4 & 1.2
& 58.6 & 80.9 & \textbf{86.9}
& 33.3 & 59.2 & \textbf{65.6} \\

Qwen2.5-7B-Instruct
& \textbf{34.2} & 2.1 & 1.2
& 57.9 & 60.8 & \textbf{80.1}
& 32.9 & \textbf{53.0} & 38.3 \\

Qwen2-72B-Instruct
& \textbf{43.5} & 9.5 & 0.9
& 71.0 & \textbf{88.6} & 84.3
& 40.0 & 51.8 & \textbf{61.6} \\

Meta-Llama-3-8B-Instruct
& \textbf{38.6} & 6.6 & 0.9
& 67.7 & 89.4 & \textbf{89.7}
& 29.0 & \textbf{57.4} & 47.0 \\

Llama-3.1-8B-Instruct
& \textbf{39.1} & 6.0 & 1.2
& 65.5 & \textbf{90.5} & 90.4
& 25.7 & \textbf{49.9} & 38.4 \\
\bottomrule
\end{tabular}
}
\vspace{-1 em}
\end{table*}

Because each molecule is evaluated under every identifier, these differences
can be tested molecule by molecule rather than only through aggregate
accuracy. Paired McNemar tests~\citep{mcnemar1947note} confirm both directions:
trivial names outperform SMILES for therapeutic-indication retrieval, whereas
SMILES outperforms trivial names for benzene detection, with both effects
significant across all seven models ($p<10^{-7}$). Exact paired counts,
$p$-values, and results for all seven benchmark tasks are reported in
Appendix~\ref{app:identifier_access_details}.
The key result is therefore not that one identifier is generally easier than
another. Instead, the identifier determines \emph{which knowledge is
accessible}: lexical names favor entity-associated knowledge, while
structure-descriptive identifiers favor information recoverable from
molecular form. This raises a key question: is the SMILES failure simply a
failure of molecular recognition?
\vspace{-1 em}
\subsection{Recognition Does Not Imply Retrieval}
\label{sec:recognition}
\vspace{-0.5 em}
To test whether failed molecular recognition explains the retrieval gap, we
evaluate two complementary recognition criteria. First, we use a
two-alternative forced-choice (2AFC) test, a standard forced-choice paradigm
in which the model must select the correct answer from two alternatives. Given a SMILES string, the model chooses between
the correct trivial name and one incorrect candidate; we retain only molecules
answered correctly across three independent decoy conditions. Second, in the
stricter free-form test, the model must generate the correct trivial name
directly from SMILES without candidate answers. Full recognition protocols and subset-construction details are
provided in Appendix~\ref{app:recognition}.

\begin{table}[t]
\vspace{-1.5 em}
\centering
\caption{Therapeutic-indication accuracy (\%) on molecules whose
SMILES--name correspondence is recognized. ``Rest'' denotes SMILES
accuracy on molecules outside the recognized subset. Higher accuracy
indicates better performance; $n$ denotes the size of the recognized subset.}
\label{tab:recognition}
\small
\setlength{\tabcolsep}{3.2pt}
\begin{tabular}{lrrrrrrrr}
\toprule
& \multicolumn{4}{c}{2AFC recognition}
& \multicolumn{4}{c}{Free-form recognition} \\
\cmidrule(lr){2-5}\cmidrule(lr){6-9}
Model & $n$ & SMILES & Name & Rest
      & $n$ & SMILES & Name & Rest \\
\midrule
ChemDFM-v2.0-14B & 576 & 4.0 & 42.7 & 2.8 & 62 & 3.2 & 40.3 & 3.6 \\
ChemDFM-v1.5-8B & 487 & 3.7 & 22.0 & 4.9 & 72 & 8.3 & 19.4 & 3.9 \\

Qwen2.5-14B-Instruct & 332 & 1.2 & 44.9 & 1.4
                  & 1 & 0.0 & 100.0 & 1.3 \\
Qwen2.5-7B-Instruct & 240 & 0.0 & 40.4 & 1.7
                  & 1 & 0.0 & 100.0 & 1.2 \\
Qwen2-72B-Instruct & 376 & 1.3 & 42.6 & 0.6
                  & 7 & 0.0 & 42.9 & 0.9 \\
Llama-3.1-8B-Instruct
& 312 & 1.6 & 42.6 & 1.4
& 6 & 0.0 & 33.3 &1.5 \\

\bottomrule
\end{tabular}
\vspace{-1.5 em}
\end{table}
As shown in Table~\ref{tab:recognition}, under 2AFC,
ChemDFM-v2.0-14B correctly recognizes 576 molecules, yet
therapeutic-indication accuracy through SMILES is only 4.0\%, close to
2.8\% on the remaining molecules, while the same recognized molecules
reach 42.7\% through their trivial names. Free-form recognition gives
the same result: among the 62 molecules whose trivial names are correctly
generated from SMILES, indication accuracy is 40.3\% through the name
but only 3.2\% through SMILES, compared with 3.6\% outside the recognized
subset. ChemDFM-v1.5-8B shows the same broader separation: under 2AFC,
the recognized molecules reach 22.0\% through their trivial names but
only 3.7\% through SMILES. The broader 2AFC pattern also holds across
the other models with recognized subsets. Recognition therefore does
not explain the retrieval gap: a model can map a structural identifier
to the correct lexical identity while access to entity-associated
knowledge remains strongly identifier-dependent. We refer to this
phenomenon as cross-identifier binding failure.
\vspace{-1.5 em}
\section{Knowledge Is Present but Unreachable}
\label{sec:unreachable}
\vspace{-1 em} 
Recognition alone does not restore factual retrieval
(Section~\ref{sec:recognition}). We therefore ask whether the missing
knowledge is absent from the model or instead inaccessible through the
structural identifier.
\vspace{-2 em}
\subsection{Lexical Identity Restores Factual Access}
\label{sec:bridge}
\vspace{-0.5 em}
We supply the missing identifier correspondence by augmenting each SMILES
with its correct trivial name, without providing any information about the
queried indication. As shown in Table~\ref{tab:bridge}, this simple
intervention nearly closes the gap to name-only retrieval across all models, recovering 91.8--99.4\% of the lost accuracy. For
ChemDFM-v2.0-14B, indication accuracy rises from 2.1\% with SMILES alone to
36.3\% with SMILES+Name, compared with 38.6\% from the name alone.
Qwen2.5-14B similarly rises from 1.3\% to 37.6\%, versus 37.8\% from the
name. Thus, supplying lexical identity is sufficient to make much of the
otherwise inaccessible factual knowledge reachable. Exact prompts, the
recovery definition, and reverse-direction controls are provided in
Appendix~\ref{app:bridge}.
\begin{table}[t]
\centering
\caption{
Supplying lexical identity restores factual access.
Therapeutic-indication accuracy (\%). Higher values indicate better performance.
}
\label{tab:bridge}
\small
\setlength{\tabcolsep}{5pt}
\begin{tabular}{lrrrr}
\toprule
Model & SMILES & + Name & Name & Recovery \\
\midrule
ChemDFM-v2.0-14B      & 2.1 & 36.3 & 38.6 & 93.6 \\
ChemDFM-v1.5-8B       & 3.9 & 21.4 & 23.0 & 91.8 \\
Qwen2.5-14B-Instruct  & 1.3 & 37.6 & 37.8 & 99.4 \\
Qwen2.5-7B-Instruct   & 1.2 & 32.8 & 33.9 & 96.6 \\
Llama-3.1-8B-Instruct & 1.2 & 38.6 & 39.0 & 99.1 \\
Qwen2-72B-Instruct    & 0.8 & 42.5 & 43.5 & 97.7 \\
\bottomrule
\end{tabular}
\vspace{- 1 em}
\end{table}
\vspace{-2 em}
\subsection{Conflicting Identifiers Expose Separate Access Routes}
\label{sec:conflict}
\vspace{-0.0 em}
To test whether lexical and structural identifiers access a single coherent
entity representation, we pair the SMILES of molecule \(A\) with the
incorrect trivial name of molecule \(B\). The resulting answers split by
query type: identity and therapeutic indication largely follow the supplied
name \(B\), whereas benzene-ring remains close to those obtained
from \(A\)'s SMILES alone. As shown in Table~\ref{tab:conflict}, for ChemDFM-v2.0-14B, the conflicting input
produces \(B\)'s identity in 85.5\% of cases and retains 32.3\% indication
accuracy for \(B\), while benzene accuracy for \(A\) remains 95.3\%, compared
with 94.5\% from its SMILES alone. Full results are provided in Appendix~\ref{app:conflict_protocol}.

\begin{table}[t]
\centering
\caption{
Conflicting identifiers control different outputs: Each conflicting prompt
pairs the SMILES of molecule A with the incorrect trivial name of molecule B.
For identity, we report how often the model returns the true molecule A or the
supplied decoy B. For therapeutic indication, we report accuracy for B under
the conflicting prompt and when B is queried by its name alone. For
benzene-ring presence, we report accuracy for A under the conflicting prompt
and when A is queried by its SMILES alone. All values are percentages.
These quantities are diagnostic and do not have a single preferred direction.
}

\label{tab:conflict}
\small
\setlength{\tabcolsep}{3pt}
\begin{tabular}{lrrrrrr}
\toprule
& \multicolumn{2}{c}{Identity under conflict}
& \multicolumn{2}{c}{Indication of B}
& \multicolumn{2}{c}{Benzene of A} \\
\cmidrule(lr){2-3}
\cmidrule(lr){4-5}
\cmidrule(lr){6-7}
Model
& True A & Decoy B
& Conflict & Name only
& Conflict & SMILES only \\
\midrule
ChemDFM-v2.0-14B
& 1.0 & 85.5
& 32.3 & 38.6
& 95.3 & 94.5 \\

ChemDFM-v1.5-8B
& 0.6 & 73.4
& 17.4 & 23.0
& 80.5 & 85.0 \\

Qwen2.5-14B-Instruct
& 0.0 & 81.1
& 35.9 & 37.8
& 84.9 & 86.9 \\

Qwen2.5-7B-Instruct
& 0.0 & 60.4
& 32.1 & 33.9
& 75.9 & 80.2 \\

Llama-3.1-8B-Instruct
& 0.0 & 86.8
& 37.3 & 39.0
& 89.3 & 90.3 \\

Qwen2-72B-Instruct
& 0.0 & 77.8
& 43.2 & 44.1
& 83.9 & 84.3 \\
\bottomrule
\end{tabular}
\vspace{-1.5 em}
\end{table}

\textbf{The conflict is often undetected.}
We next ask whether models recognize that the two supplied identifiers are
incompatible. When explicitly asked whether a SMILES and trivial name refer
to the same molecule, models reject obvious mismatches much more reliably
than structurally similar ones. Across the tested models, only 17.5--43.1\%
of similar incorrect names are rejected; even in a two-choice identification
test, accuracy is only 56.4--73.9\%. Thus, conflicting identifiers can drive
different answers without the model reliably detecting the inconsistency.
Full prompts, decoy construction, and discrimination results are reported in
Appendix~\ref{app:conflict_detection}.
\vspace{-1 em}
\section{Diagnosing and Restoring Cross-Identifier Access}
\label{sec:diagnosis}
\vspace{-1 em}
\subsection{Causal Analysis via Activation Patching}
\label{sec:patching}
\vspace{-0.5 em}
The behavioral results suggest that factual knowledge is accessible through a
molecule's trivial name but not through its SMILES. We test whether this
difference can be causally traced to the internal state induced by the
identifier. We use \emph{activation patching}~\citep{zhang2024towards}, a causal intervention that replaces an
internal activation in a target computation with the corresponding activation
from a source computation, to test whether the lexical state is sufficient to
restore factual retrieval.
For matched prompts querying the same fact, we treat the trivial-name run as
the source and the SMILES run as the target. At each tested layer, we replace
the residual-stream state at the final prompt token of the SMILES run with the
corresponding source state.
We measure recovery as the fraction of the name--SMILES teacher-forced
log-probability gap closed by the intervention; a different-molecule activation
with a different gold answer serves as the control. Full prompts, scoring, and implementation details are provided in Appendix~\ref{app:patching}.

\begin{table}[t]
\centering
\caption{
Lexical activations causally restore retrieval from SMILES. Peak recovery (\%)
of the name--SMILES log-probability gap and the corresponding relative layer
depth. Control patches use a different molecule with a different gold answer.
Higher recovery indicates stronger restoration; peak depth is descriptive,
while the control is expected not to yield positive recovery.
}
\label{tab:activation_patching}
\small
\setlength{\tabcolsep}{4.2pt}
\begin{tabular}{lrrr}
\toprule
Model & Peak depth & Recovery & Control \\
\midrule
ChemDFM-v2.0-14B       & 0.71 & 105.4 & -88.3 \\
ChemDFM-v1.5-8B        & 0.50 &  76.4 & -79.6 \\
Qwen2.5-14B-Instruct   & 0.71 & 106.0 & -197.7 \\
Qwen2.5-7B-Instruct    & 0.75 & 104.9 & -111.4 \\
Meta-Llama-3-8B-Instruct
                       & 0.52 & 135.0 & -179.9 \\
Llama-3.1-8B-Instruct  & 0.50 & 112.4 & -208.2 \\
Qwen2-72B-Instruct     & 0.83 & 138.4 & -30.9 \\
\bottomrule
\end{tabular}
\vspace{-2 em}
\end{table}

As shown in Table~\ref{tab:activation_patching}, the correct lexical state
substantially restores factual retrieval in every model, with peak recovery
ranging from 76.4\% to 138.4\%. In contrast, every different-molecule control
is negative. The effect peaks at relative depths 0.50--0.83, rather than at a
single architecture-specific layer. These interventions provide causal
evidence that the lexical route produces a retrieval-supporting state that
the SMILES route fails to construct on its own. We next ask whether this
access asymmetry is reflected in the geometry of the two identifier
representations.
\vspace{-1 em}
\subsection{Probing Cross-Identifier Binding in Model Representations}
\label{sec:representations}
\vspace{-0.5 em}

We examine whether co-referring molecular identifiers become aligned at the
level of individual entities. If two identifier representations encode the same
molecular identity in a shared geometry, the representation of a molecule under
one identifier should retrieve its co-referring representation under another
identifier from a large candidate set. We therefore measure cross-identifier
top-1 retrieval.
For each of the 896 molecules, we extract representations from
identifier-specific prompts and use the representation under one identifier to
retrieve the corresponding SMILES representation from all 896 candidates.
We evaluate retrieval across hidden-state depths and report the maximum top-1
accuracy. Throughout this section, Name--SMILES denotes trivial
name $\rightarrow$ SMILES retrieval, whereas IUPAC--SMILES denotes IUPAC
name $\rightarrow$ SMILES retrieval. 
Full representation extraction, pooling, centering, normalization, and layer-wise
evaluation details are provided in Appendix~\ref{app:representation}.
\begin{figure}[t]
    \centering
    \includegraphics[width=\linewidth]{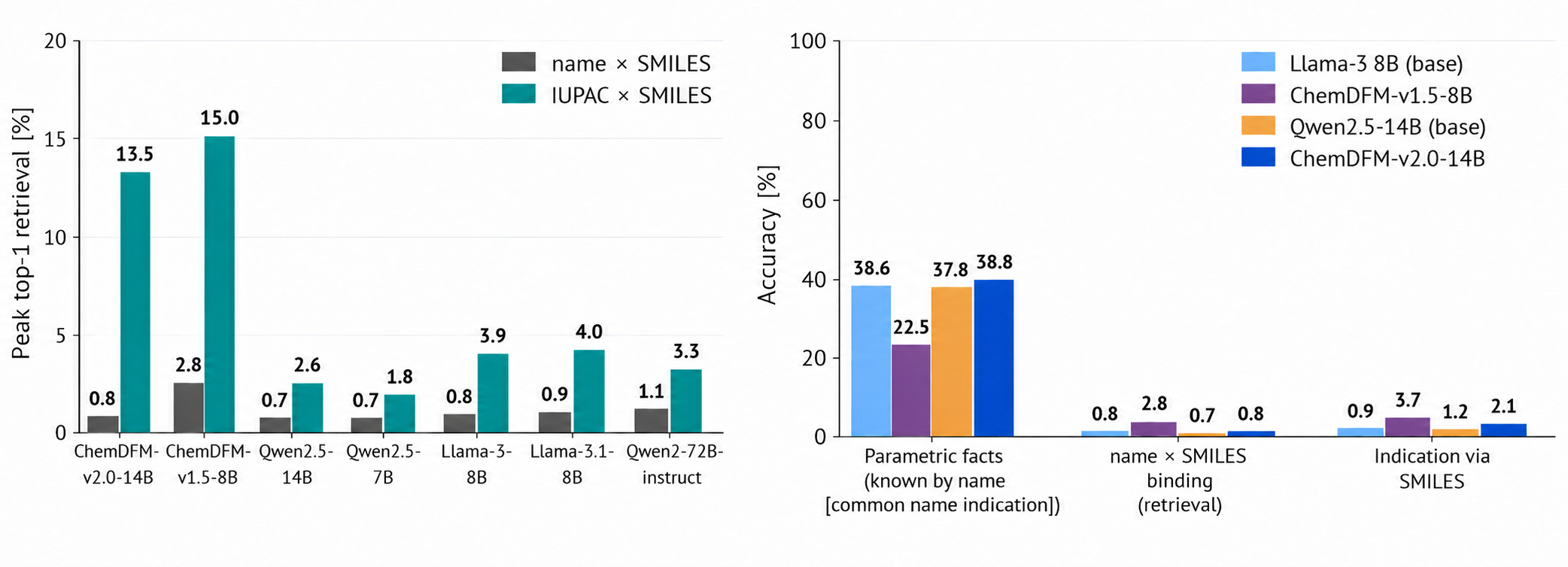}
    \vspace{-3 em}
    \caption{\textbf{Cross-identifier representation alignment and chemistry
    specialization.}
    \textbf{Left:} peak top-1 retrieval of the same molecule among 896
    candidates across hidden-state depths; chance performance is \(0.11\%\).
    Name--SMILES denotes trivial name \(\rightarrow\) SMILES retrieval and
    IUPAC--SMILES denotes IUPAC name \(\rightarrow\) SMILES retrieval.
    \textbf{Right:} comparison of the two general-purpose reference models
    with their chemistry-specialized counterparts. ``Parametric facts known by
    name'' denotes strict-scored trivial-name indication accuracy;
    ``Name--SMILES binding'' denotes the same cross-identifier retrieval
    metric; and ``Indication via SMILES'' denotes strict-scored
    SMILES-conditioned indication accuracy.}
    \label{fig:representation}
    \vspace{-1.5 em}
\end{figure}
The result reveals a sharp asymmetry. Even at the best depth, Name--SMILES
retrieval reaches only \(0.7\)--\(2.8\%\) across all models. Yet weak alignment is not inevitable across molecular
identifiers: IUPAC--SMILES retrieval reaches \(13.5\%\) for
ChemDFM-v2.0-14B and \(15.0\%\) for ChemDFM-v1.5-8B
(Figure~\ref{fig:representation}). The same models can therefore develop
substantially stronger correspondence between two structure-descriptive forms
while leaving lexical identity only weakly connected to molecular structure.

Chemistry specialization makes this separation especially clear. For the
Meta-Llama-3-8B-Instruct \(\rightarrow\) ChemDFM-v1.5-8B comparison,
IUPAC--SMILES retrieval rises from \(3.9\%\) to \(15.0\%\), whereas
Name--SMILES retrieval rises only from \(0.8\%\) to \(2.8\%\). For
Qwen2.5-14B-Instruct \(\rightarrow\) ChemDFM-v2.0-14B,
IUPAC--SMILES retrieval rises from \(2.6\%\) to \(13.5\%\), while
Name--SMILES retrieval changes only from \(0.7\%\) to \(0.8\%\). Thus, the chemistry-specialized models exhibit substantially stronger alignment between the two structure-descriptive identifiers than their general-purpose reference models
without producing comparable
binding between molecular structure and lexical identity. This representational asymmetry mirrors the behavioral access gap. Stronger
chemical representations do not by themselves make entity-associated factual
knowledge reliably accessible through SMILES: the models can substantially
align IUPAC names with molecular structure while leaving the lexical route
through which much of their factual knowledge is accessed only weakly
connected to that same structure.
\vspace{-0.5 em}
\vspace{-0.5 em}
\subsection{Lexical Routing Restores Factual Access from SMILES}
\label{sec:lookup}
\vspace{-0.5 em}
The lexical bridge above suggests a simple inference mitigation: resolve a
SMILES string to its lexical identity before asking for the fact. We test this
over 891 molecules and 2,955 available molecule--relation pairs spanning
therapeutic indication, molecular target, mechanism of action, and ATC class.
For each query, we compare direct SMILES access, SMILES augmented with the
molecule's ChEMBL preferred name obtained by database lookup, and the preferred
name alone. The lookup supplies only the identifier correspondence, not the
queried fact. Full construction, prompts, scoring, and controls are given in
Appendix~\ref{app:resolver}.

\begin{table}[t]
\centering
\vspace{0.45 em}
\caption{
Aggregate accuracy (\%) over available molecule--relation pairs.
Higher values indicate better performance.
}
\label{tab:lookup}
\small
\setlength{\tabcolsep}{4pt}
\begin{tabular}{lrrrr}
\toprule
Model & SMILES & + Lookup & Name & Gap rec. \\
\midrule
ChemDFM-v2.0-14B         & 2.4 & 20.8 & 24.4 & 83.6 \\
ChemDFM-v1.5-8B          & 5.4 & 14.2 & 14.5 & 96.7 \\
Qwen2.5-14B-Instruct     & 1.2 & 25.3 & 29.2 & 86.1 \\
Qwen2.5-7B-Instruct      & 0.6 & 22.6 & 23.2 & 97.3 \\
Meta-Llama-3-8B-Instruct & 1.2 & 26.8 & 27.1 & 98.8 \\
Llama-3.1-8B-Instruct    & 0.6 & 26.8 & 28.9 & 92.6 \\
\bottomrule
\end{tabular}
\vspace{-1 em}
\end{table}

As Table~\ref{tab:lookup} shows, lexical routing closes
\(83.6\)--\(98.8\%\) of the aggregate SMILES--name gap across all models. The model itself rarely supplies this bridge reliably: in a
self-resolution control, SMILES-to-name identification succeeds for only
\(0.1\)--\(12.0\%\) of molecules (Appendix~\ref{app:resolver}). Thus, most of
the missing factual access can be recovered by supplying the identifier
correspondence, but that correspondence is not reliably inferred from SMILES
alone.
Lookup-based routing therefore provides an effective inference workaround,
but it requires lexical resolution on every query and remains bounded by the
knowledge accessible through the name. Can the same access
instead be written into the model parameters, using standard knowledge editors
as a diagnostic of what is missing?
\vspace{-1 em}
\subsection{Standard Editors Do Not Reliably Transfer Across Identifiers}
\label{sec:editing}
\vspace{-0.5 em}
Can the lookup required above be avoided by modifying the model weights?
We test this diagnostically with ROME~\citep{gupta-etal-2024-unified,gupta-etal-2024-model} and fine-tuning (FT)~\citep{ouyang2022training}: a counterfactual
drug indication is edited through the molecule's trivial name and then queried
through its co-referring SMILES. Because an edit may also increase production
of the target for unrelated molecules, we subtract the corresponding locality
leakage,
\(T_{\mathrm{specific}}=T_{\mathrm{SMILES}}-T_{\mathrm{leak}}\).
Full edit construction, IUPAC results, layer selection, and hyperparameters are
given in Appendix~\ref{app:editing}.
\begin{table}[t]
\centering
\caption{
All values are percentages. Specific transfer is computed as SMILES transfer
minus leakage and is therefore reported in percentage points; brackets denote
95\% bootstrap confidence intervals. Higher edit efficacy, specific transfer,
and +Name recovery indicate better performance, while lower leakage is better.
Raw SMILES transfer should be interpreted together with leakage.
}
\label{tab:editing}
\small
\setlength{\tabcolsep}{3.2pt}
\begin{tabular}{llrrrrr}
\toprule
Model & Method & Effic. & SMILES & Leak & Specific [95\% CI] & +Name \\
\midrule
ChemDFM-v2.0-14B
& ROME & 52.2 & 22.1 & 19.8 & +2.3 [-0.9,5.5] & 45.5 \\
& FT & 26.0 & 1.8 & 2.1 & -0.3 [-1.6,1.2] & 25.2 \\
Qwen2.5-14B
& ROME & 17.9 & 4.4 & 4.9 & -0.5 [-2.4,1.4] & 12.3 \\
& FT & 29.0 & 11.5 & 8.0 & \textbf{+3.5 [0.9,6.3]} & 26.7 \\
Qwen2.5-7B
& ROME & 47.9 & 4.0 & 7.4 & -3.4 [-5.7,-0.9] & 11.4 \\
& FT & 34.5 & 5.4 & 7.3 & -1.9 [-4.4,0.8] & 21.7 \\
Llama-3.1-8B
& ROME & 44.2 & 21.1 & 8.1 & \textbf{+13.0 [9.2,16.9]} & 40.4 \\
& FT & 38.3 & 12.6 & 11.4 & +1.2 [-1.6,4.1] & 32.1 \\
\bottomrule
\end{tabular}
\vspace{-1 em}
\end{table}
Across the eight model--editor configurations, six show no reliable
entity-specific transfer to SMILES beyond the leakage baseline, while only two exhibit statistically reliable transfer beyond this baseline, including substantial transfer for
Llama-3.1-8B-Instruct with ROME (Table~\ref{tab:editing}). This pattern remains
when we restrict evaluation to edits that demonstrably succeed through the
trivial name (Appendix~\ref{app:editing}), showing that weak transfer is not
simply caused by failed edits. Moreover, restoring the correct name alongside
SMILES often makes the edited association accessible again. Thus, standard
editing can sometimes propagate across identifiers, but this behavior is
highly inconsistent across models and methods: modifying the factual
association does not reliably repair the structural route used to access it.

\vspace{-1 em}
\section{Binding Self-Distillation Transfers Lexical Access to SMILES}
\label{sec:binding}
\vspace{-1 em}
The editing results suggest that changing a factual association is not enough:
the missing object is the \emph{access route} from SMILES to knowledge that is
already reachable through the molecule's lexical identity. We therefore ask
whether this access behavior itself can be written into the model. Our goal is
not to teach a new fact, but to make the SMILES route reproduce knowledge
access that the same model already exhibits when the correct name is available. We introduce \textbf{Binding Self-Distillation} (BSD). For a molecule with SMILES
\(s\), trivial name \(n\), and a generic continuation stem \(z\), we construct
matched teacher and student contexts,
\[
x_T=[\mathrm{SMILES}(s);\text{ ``known as }n\text{''};z],
\qquad
x_S=[\mathrm{SMILES}(s);z].
\]
Teacher and student use the same pretrained model. The teacher is frozen and
receives the lexical identity, while the student receives only SMILES and
learns a lightweight LoRA adapter by minimizing $
D_{\mathrm{KL}}
\bigl(
p_{\theta}(\cdot\mid x_T)
\;\|\;
p_{\theta,\phi}(\cdot\mid x_S)
\bigr).
$
Thus, the training signal is the model's own name-conditioned predictive
behavior: no factual labels, generated target answers, or external teacher are
used. We train one adapter per molecule using generic continuation stems, with
the downstream factual evaluation prompts excluded from training. At
inference time, the name is removed and only SMILES is provided. Full training configuration, prompts, and controls are
in Appendix~\ref{app:binding}.
\begin{table*}[t]
\centering
\caption{
Facts/entity is the average number of available factual relations
retrieved per molecule before and after the per-entity binding intervention.
Higher Facts/entity indicates better performance; $n$ is the model-specific number of molecules retained by the original BSD entity-selection pipeline.
}

\label{tab:binding}
\small
\setlength{\tabcolsep}{5pt}
\begin{tabular}{lrr}
\toprule
Model & \(n\) & Facts/entity \\
\midrule
ChemDFM-v2.0-14B       & 389 & \(0.05 \rightarrow 0.87\) \\
Qwen2.5-14B-Instruct   & 390 & \(0.06 \rightarrow 0.73\) \\
Llama-3.1-8B-Instruct  & 389 & \(0.03 \rightarrow 0.68\) \\
ChemDFM-v1.5-8B        & 47  & \(0.51 \rightarrow 0.85\) \\
Qwen2.5-7B-Instruct    & 351 & \(0.04 \rightarrow 0.85\) \\
Meta-Llama-3-8B-Instruct & 378 & \(0.02 \rightarrow 0.78\) \\
Qwen2-72B-Instruct     & 435 & \(0.01 \rightarrow 0.85\) \\
\bottomrule
\end{tabular}
\vspace{-1.5 em}
\end{table*}
Binding Self-Distillation increases factual access through SMILES across all
seven models (Table~\ref{tab:binding}). For ChemDFM-v2.0-14B, the number of
retrieved facts rises from \(0.05\) to \(0.87\) per entity. 
The gain is not restricted to the indication relation used to select eligible
molecules: SMILES-conditioned accuracy rises from 2.1\% to 51.4\% for
indication, 0.0\% to 27.6\% for molecular target, 0.0\% to 5.0\% for
mechanism of action, and 3.3\% to 11.4\% for ATC class
(Appendix~\ref{app:binding_all_models},
Table~\ref{tab:binding_crossmodel_full}).
The effect also
persists at substantially larger scale: Qwen2-72B-Instruct rises from
\(0.01\) to \(0.85\) facts per entity over 435 molecules. Because the factual
evaluation prompts are excluded from training, these improvements reflect
transfer of the lexical access behavior across prompts rather than direct
training on the evaluation answers.

How much of the lexical route can be internalized? On the same
ChemDFM-v2.0-14B entities, explicitly supplying the correct name raises access
from 0.05 to 1.01 facts per entity, while Binding Self-Distillation reaches
0.87, recovering 86\% of the available SMILES--name gap without requiring the
name at inference time. For Qwen2-72B-Instruct, the corresponding values are
0.01, 1.34, and 0.85, yielding 63\% recovery
(Appendix~\ref{app:binding_bridge},
Table~\ref{tab:binding_bridge_full}). Retention controls for these two
highlighted settings also show that the intervention need not destroy behavior
that already works: benzene verdicts are preserved for 94.1\% of
ChemDFM-v2.0-14B examples and 92.9\% of Qwen2-72B-Instruct examples, with
additional locality analyses reported in
Appendix~\ref{app:binding_retention}.
Together, these results provide a constructive complement to the editing
diagnostic in Section~\ref{sec:editing}. Standard editors modify factual
associations but propagate them inconsistently across identifiers; Binding
Self-Distillation instead targets the missing access relation. The result
shows that knowledge already exposed by a lexical identifier can be made
substantially more accessible from SMILES without requiring that lexical
identifier at inference time.

\vspace{-1 em}
\section{Conclusion}
\label{sec:discussion}
\vspace{-1 em}
In chemistry pipelines, molecular identity is often treated as a solved
interface problem: once two identifiers are known to denote the same molecule,
an LLM is implicitly expected to access the same underlying knowledge. We
introduce \textbf{ChemIDBench}, a controlled benchmark for evaluating knowledge
access across molecular representations, and find that across LLMs spanning
7B to 72B, changing only the identifier can change which knowledge is
reachable. Surprisingly, a model can identify a molecule from SMILES and
retrieve a fact through its trivial name, yet fail to retrieve the same fact
directly from SMILES. When identifiers conflict, different routes can also
drive different answers within the same prompt, often without the model
detecting the inconsistency. Recognition controls, conflict experiments,
representation analysis, and activation patching together show that entity
recognition, knowledge possession, and identifier-conditioned access are
separable capabilities. We call this \textbf{cross-identifier binding failure}.
Standard knowledge editors do not reliably restore access across identifiers;
we therefore introduce \textbf{Binding Self-Distillation}, which directly
targets the missing access relation and substantially improves factual retrieval
from SMILES without requiring the lexical name at inference time. Our current
formulation performs this repair per entity, showing that cross-identifier
access is repairable while leaving generalization to unseen entities as a
future direction. Beyond chemistry, the broader identifier-dependent-access
phenomenon also appears in our protein-domain replication with opaque UniProt
accessions~\citep{ahmad2025uniprot} in Appendix~\ref{app:biology}. Overall, our
results show that for LLMs to be used in science, knowledge is only useful if
it remains accessible across the different representations of the same entity.

\bibliography{iclr2027_conference}
\bibliographystyle{iclr2027_conference}

\newpage
\appendix

\section{Related Work} \label{RL}
\paragraph{LLMs and molecular representations in Chemistry.}
LLMs have increasingly been adapted to chemistry through domain-specific
pretraining and instruction tuning. ChemLLM~\citep{zhang2024chemllm} and
ChemDFM~\citep{zhao2025developing}, for example, train language models on
chemical knowledge and tasks that include textual molecular representations
such as SMILES~\citep{gomez2018automatic,schwaller2019molecular}. In practice, the same molecule may be presented to these
models through a trivial ~\citep{gomez2018automatic,schwaller2019molecular}, a systematic IUPAC name~\citep{gomez2018automatic,schwaller2019molecular}, or a SMILES string~\citep{gomez2018automatic,schwaller2019molecular}.
Our work asks whether these co-referring representations are interchangeable
not simply as inputs for downstream task performance, but as access routes to
knowledge already stored in the model.
\vspace{-1 em}
\paragraph{Representation dependence and consistency in chemistry.}
Recent work has shown that molecular representation can substantially affect
LLM behavior. \citet{baker-etal-2025-molecular} compare multiple molecular
representations for property prediction and find strong representation
preferences across pretrained LLMs. \citet{yan2025inconsistency} study
cross-representation consistency between chemically equivalent SMILES and
IUPAC inputs, showing that equivalent representations can produce inconsistent
predictions and that consistency regularization only partially reduces this
gap. Even within SMILES, alternative valid encodings of the same molecule can
change learned representations and downstream performance~\citep{kikuchi2025notation}. These studies establish that chemically
equivalent representations are not behaviorally interchangeable. We ask a
different question: whether co-referring identifiers provide access to the
\emph{same knowledge already present in the model}. We distinguish failed
recognition from failed retrieval, separate entity-associated facts from
structure-computable properties, and examine what happens when two individually
valid but mutually inconsistent identifiers are presented together.
\vspace{-1 em}
\paragraph{Mechanistic analysis of molecular and factual representations.}
Recent mechanistic work has examined what chemical language models encode in
their internal representations~\citep{cohen2025unveiling,varadi2025circuits,kenneth2026does,licontrollable,zhang2026slim}. Sparse-autoencoder analyses have identified
latent features associated with molecular syntax, structural motifs,
physicochemical properties, and pharmacological concepts
~\citep{cohen2025unveiling,kenneth2026does}, while layerwise probing and causal
interventions have been used to distinguish information that is merely
decodable from information that is functionally used by molecular models
~\citep{hao2026mechanistic,varadi2025circuits}. These studies primarily ask
what molecular structure and properties are represented inside models trained
on molecular strings. Our question is different: why can knowledge about the
same entity be accessible through one co-referring identifier but not another?
We therefore study not only whether information is represented, but whether
identifier-specific internal states provide effective access to the same
factual knowledge.
\vspace{-1 em}
\paragraph{Knowledge editing.}
Knowledge editing methods aim to modify selected factual associations while
preserving unrelated model behavior. ROME and fine-tuning
~\citep{gupta-etal-2024-unified,gupta-etal-2024-model} provide representative
parameter-based approaches, while subsequent work has broadened evaluation
beyond edit success and paraphrase generalization. Recent evaluations
~\citep{hsueh-etal-2024-editing,jiang-etal-2024-learning} further test whether
an edit propagates to logically related facts. In chemistry,
MolEdit~\citep{lei2026moledit} considers knowledge editing in multimodal
molecule--language models and evaluates reliability, locality, and generality
for molecule--text generation tasks. We introduce a different generalization
requirement: \emph{cross-identifier transfer}. If two identifiers denote the
same entity, an edit made through one identifier should remain accessible when
that entity is queried through another. We find that standard editors do not
reliably satisfy this requirement for trivial-name--to--SMILES transfer,
motivating cross-identifier binding as a distinct intervention target from
conventional edit efficacy, paraphrase generalization, or cross-modal editing.

\section{Benchmarking Cross-Identifier Knowledge Access:
Additional Details}
\label{app:benchmark}

\subsection{Benchmark Design}
\label{app:benchmark_details}

\paragraph{Initial ChEMBL retrieval.}
We select ChEMBL~\citep{mendez2019chembl} drug-indication records corresponding
to approved indications (phase 4) and group them by molecule. For each molecule,
we retain its ChEMBL preferred name, canonical SMILES, and all qualifying MeSH
indication headings. This produces 3,025 unique molecules before subsequent
entity and structure filtering.

\paragraph{Entity and structure filtering.}
Because our benchmark compares different identifiers for the \emph{same}
molecular entity, we first restrict the collection to molecules with
well-defined lexical and structural representations. We retain only candidates
with both a ChEMBL preferred name and canonical SMILES, and restrict preferred
names to at most two whitespace-separated words to exclude long,
formulation-like labels. Preferred names are stripped and lowercased before
filtering. To focus on single molecular entities rather than salts,
counterions, or mixtures, we remove salt and counterion forms by exact
whitespace-token matching against a fixed manually specified
lexicon,\footnote{The lexicon contains
\texttt{sulfate}, \texttt{sulphate}, \texttt{hydrochloride},
\texttt{chloride}, \texttt{sodium}, \texttt{potassium}, \texttt{calcium},
\texttt{magnesium}, \texttt{acetate}, \texttt{citrate}, \texttt{tartrate},
\texttt{maleate}, \texttt{mesylate}, \texttt{besylate}, \texttt{fumarate},
\texttt{succinate}, \texttt{phosphate}, \texttt{nitrate}, \texttt{bromide},
\texttt{iodide}, \texttt{carbonate}, \texttt{gluconate}, \texttt{lactate},
\texttt{oxalate}, \texttt{salicylate}, \texttt{stearate}, \texttt{tosylate},
\texttt{dihydrate}, \texttt{monohydrate}, \texttt{hydrate},
\texttt{hydrobromide}, \texttt{bitartrate}, \texttt{pamoate}, and
\texttt{valerate}.}
and exclude multi-fragment structures whose ChEMBL canonical SMILES contains
``.''.

We further remove duplicate preferred names and structures so that an entity is
not represented multiple times in the benchmark. Duplicate preferred names are
identified by exact equality after lowercasing, retaining the first molecule in
sorted ChEMBL-ID order. Remaining SMILES are parsed with
RDKit~\citep{rdkit}, ensuring that structure-derived ground truth can be
computed reliably, and we restrict the parsed molecular graphs to 6--70 atoms
to bound the benchmark to a consistent molecular-size range. For structure
deduplication, each parsed molecule is canonicalized as an isomeric SMILES and
duplicate structures are removed by exact equality of the resulting strings.
These filters leave 1,553 candidate molecules.

\paragraph{Therapeutic-indication target construction.}
For each retained molecule, we collect all MeSH headings associated with
phase-4 drug indications in ChEMBL. To make answer matching robust to
superficial formatting differences, we normalize each heading by lowercasing,
replacing non-alphanumeric characters with spaces, and collapsing repeated
whitespace. For headings containing commas, we additionally include a
comma-inverted variant to accommodate alternative word orderings of the same
MeSH heading. Duplicate variants are removed, and strings of length at most
three characters are discarded to avoid extremely short, non-discriminative
matches.

We then compute document frequency over the accepted aliases at the molecule
level, counting each alias at most once per molecule, and remove aliases
occurring in more than 2\% of the 1,553 candidates. This excludes overly
generic indication terms that provide little molecule-specific signal. Because
a molecule may have multiple qualifying MeSH headings, we define a single
deterministic reference indication as the lexicographically first unique
original heading; this is only a reproducible tie-breaking rule and does not
imply clinical priority. A molecule is retained only if at least one accepted
indication alias remains and its reference indication is still scorable under
the resulting alias set, ensuring that every retained molecule has a valid
target for the indication task. This filtering leaves 897 molecules.

\paragraph{IUPAC-name resolution.}
For each of the 897 remaining molecules, we retrieve an IUPAC name from
PubChem~\citep{kim2023pubchem} using the PubChem PUG REST interface. We first
query PubChem by the ChEMBL preferred name and take the first returned IUPAC
name. If this lookup does not yield an IUPAC name, we query PubChem for
compound identifiers associated with the preferred name, select the first
returned compound, and retrieve its IUPAC name. Only if both name-based
procedures fail do we query PubChem using the molecule's canonical SMILES.
PubChem mappings are additionally validated against the corresponding ChEMBL
structures using the InChIKey connectivity block. An IUPAC name is resolved
for 896 of the 897 candidates; the single unresolved molecule is discarded,
yielding a final benchmark of 896 molecules.

\begin{table}[t]
\centering
\caption{\textbf{Benchmark construction.} Successive filtering reduces the
initial ChEMBL collection to 896 molecules with trivial-name, IUPAC-name, and
SMILES identifier fields and scorable therapeutic-indication annotations.}
\label{tab:benchmark_construction}
\begin{tabular}{lr}
\toprule
Construction stage & \# Molecules \\
\midrule
ChEMBL molecules with qualifying phase-4 indications & 3,025 \\
Entity and structure filtering & 1,553 \\
Therapeutic-indication normalization and filtering & 897 \\
Successful PubChem IUPAC-name resolution & 896 \\
\bottomrule
\end{tabular}
\end{table}

\subsection{Evaluation Protocol}
\label{app:evaluation_protocol1}

\paragraph{Prompt construction.}
Each molecule is evaluated independently under three identifier conditions:
trivial name, IUPAC name, and canonical SMILES. For a given molecule and task,
the queried property and target answer are held fixed; only the identifier
used to refer to the molecule changes. Table~\ref{tab:identifier_prefixes}
shows the identifier-specific prefixes. The same task instruction is appended
under each identifier condition, yielding seven tasks per identifier and 21
prompts per molecule. Across 896 molecules, this results in 18,816 prompts per
model.

\begin{table}[t]
\centering
\caption{\textbf{Identifier-specific prompt prefixes.} Only the molecular
identifier changes across the three conditions.}
\label{tab:identifier_prefixes}
\begin{tabular}{lp{0.68\linewidth}}
\toprule
Identifier & Prompt prefix \\
\midrule
Trivial name &
Consider the molecule commonly known as ``\{NAME\}''. \\
IUPAC name &
Consider the molecule named \{IUPAC\}. \\
SMILES &
Consider the molecule with SMILES \{SMILES\}. \\
\bottomrule
\end{tabular}
\end{table}

\paragraph{Benchmark tasks.}
The benchmark contains two entity-associated tasks and five
structure-computable tasks. \textsc{Identify} asks the model to recover the
molecule's trivial name, while \textsc{Indication} asks for a therapeutic
indication associated with the molecule. The remaining tasks ask for molecular
formula, benzene-ring presence, nitrogen-atom count, ring count, and
heavy-atom count. Exact task instructions are provided in
Table~\ref{tab:benchmark_prompts}.

\begin{table}[t]
\centering
\caption{\textbf{Benchmark task prompts.} Each task instruction is appended
to one of the three identifier-specific prefixes in
Table~\ref{tab:identifier_prefixes}.}
\label{tab:benchmark_prompts}
\begin{tabular}{lp{0.72\linewidth}}
\toprule
Task & Prompt \\
\midrule

\textsc{Identify} &
What is the common generic name of this molecule?
Answer in English. Reply with the name only. \\

\textsc{Indication} &
What medical condition is this substance primarily used to treat?
Answer in English. Reply with the condition only. Do not explain. \\

\textsc{Formula} &
What is the molecular formula of this molecule?
Answer in English. Reply with the formula only, e.g.\ C9H8O4.
Do not explain. \\

\textsc{Benzene} &
Does this molecule contain a benzene ring?
Answer in English. Reply with exactly one word: yes or no.
Do not explain. \\

\textsc{Nitrogens} &
How many nitrogen atoms does this molecule contain?
Answer in English. Reply with a single integer only. Do not explain. \\

\textsc{Ring Count} &
How many rings does this molecule contain?
Answer in English. Reply with a single integer only. Do not explain. \\

\textsc{Heavy Atoms} &
How many non-hydrogen atoms does this molecule contain?
Answer in English. Reply with a single integer only. Do not explain. \\

\bottomrule
\end{tabular}
\end{table}

\paragraph{Ground-truth construction.}
For \textsc{Identify}, the ChEMBL preferred name is used as the gold answer.
For \textsc{Indication}, predictions are evaluated against the strict set of
accepted MeSH-derived aliases constructed as described in
Appendix~\ref{app:benchmark_details}. Ground truth for the five
structure-computable tasks is derived from the canonical SMILES using
RDKit~\citep{rdkit}. Molecular formula is calculated from the parsed molecular
structure. Benzene-ring presence is determined by testing for an aromatic
six-membered carbon ring. Nitrogen count is the number of nitrogen atoms in
the molecule, ring count is the number of rings returned by RDKit's
ring-information representation, and heavy-atom count is the number of
non-hydrogen atoms.

\paragraph{Decoding.}
All models are evaluated using greedy decoding with a maximum of 40 generated
tokens per query. Sampling is disabled, and input sequences are truncated to
at most 1,536 tokens. The same decoding procedure is used across identifier
forms and benchmark tasks.

\paragraph{Entity-associated scoring.}
For \textsc{Identify}, generated responses and ChEMBL preferred names are
normalized before matching. Normalization lowercases text, removes possessive
suffixes and punctuation, collapses repeated whitespace, and applies a light
plural normalization that removes a terminal ``s'' from tokens longer than
four characters unless they end in ``ss''. A response is counted as correct
when the normalized preferred name occurs as a complete
word-boundary-matched phrase in the normalized response.
For \textsc{Indication}, empty responses and refusals are rejected, as are responses containing more than 12 normalized tokens. Remaining responses are normalized using the same text-processing procedure and counted as correct when any alias in the strict accepted-answer set occurs as a complete word-boundary-matched phrase in the normalized response.

\paragraph{Structure-computable scoring.}
For \textsc{Formula}, generated responses are parsed into element-count
compositions after normalizing formatting variations including subscripts,
letter case, and punctuation. The scorer tests up to the first four
formula-like candidates in the response and, if none matches, parses the
complete response. A prediction is counted as correct when the resulting
element counts exactly match the RDKit-derived gold formula.

For the integer-valued tasks, \textsc{Nitrogens}, \textsc{Ring Count}, and
\textsc{Heavy Atoms}, the first integer appearing in the generated response
is extracted and compared with the corresponding RDKit-derived gold value. A
response containing no integer is counted as incorrect.

For \textsc{Benzene}, responses are interpreted using a negation-first yes/no
rule: explicit negative cues are evaluated before positive cues, ensuring that
statements such as ``does not contain'' are classified as negative. Empty,
refusal, or otherwise unparseable responses are counted as incorrect.

\subsection{Cross-Identifier Additional Results and Statistical Tests}
\label{app:identifier_access_details}

This section reports the complete cross-identifier results and paired
statistical tests supporting Section~\ref{sec:identifier_access}. Benchmark
construction and evaluation procedures are described in
Appendices~\ref{app:benchmark_details} and
\ref{app:evaluation_protocol1}, respectively.

\paragraph{Primary cross-identifier results.}
Table~\ref{tab:full_identifier_results} reports the four tasks most directly
used in our analysis. Therapeutic-indication retrieval consistently favors
the trivial name, whereas benzene-ring detection and nitrogen counting favor
structure-descriptive identifiers across models. Identification measures whether the model can recover the trivial name from IUPAC or SMILES; the trivial-name condition serves as a ceiling control because the requested name is already present in the prompt.

\begin{table*}[t]
\centering
\small
\caption{
\textbf{Primary cross-identifier results.}
Accuracy (\%) when the underlying molecule and queried property are held fixed
and only the supplied identifier changes.
\textit{Identify} denotes trivial-name recovery,
\textit{Therapeutic indication} denotes indication retrieval,
\textit{Benzene} denotes benzene-ring detection, and
\textit{Nitrogens} denotes nitrogen-atom counting.
Best performance within each model and task is shown in bold.
}
\label{tab:full_identifier_results}
\resizebox{\textwidth}{!}{
\begin{tabular}{llcccc}
\toprule
\textbf{Model} & \textbf{Identifier}
& \textbf{Identify}
& \textbf{Therapeutic indication}
& \textbf{Benzene}
& \textbf{Nitrogens} \\
\midrule

\multirow{3}{*}{ChemDFM-v2.0-14B}
& Trivial name & \textbf{93.3} & \textbf{38.8} & 76.7 & 41.6 \\
& IUPAC name   & 3.8 & 2.2 & 93.2 & \textbf{70.3} \\
& SMILES       & 6.9 & 2.1 & \textbf{94.6} & 59.6 \\
\midrule

\multirow{3}{*}{ChemDFM-v1.5-8B}
& Trivial name & \textbf{72.7} & \textbf{22.5} & 70.5 & 5.5 \\
& IUPAC name   & 2.7 & 0.1 & 81.1 & 8.7 \\
& SMILES       & 8.0 & 3.7 & \textbf{84.4} & \textbf{21.2} \\
\midrule

\multirow{3}{*}{Qwen2.5-14B-Instruct}
& Trivial name & \textbf{85.4} & \textbf{37.8} & 58.6 & 33.3 \\
& IUPAC name   & 2.9 & 5.4 & 80.9 & 59.2 \\
& SMILES       & 0.1 & 1.2 & \textbf{86.9} & \textbf{65.6} \\
\midrule

\multirow{3}{*}{Qwen2.5-7B-Instruct}
& Trivial name & \textbf{41.3} & \textbf{34.2} & 57.9 & 32.9 \\
& IUPAC name   & 1.8 & 2.1 & 60.8 & \textbf{53.0} \\
& SMILES       & 0.1 & 1.2 & \textbf{80.1} & 38.3 \\
\midrule

\multirow{3}{*}{Qwen2-72B-Instruct}
& Trivial name & \textbf{62.1} & \textbf{43.5} & 71.0 & 40.0 \\
& IUPAC name   & 5.2 & 9.5 & \textbf{88.6} & 51.8 \\
& SMILES       & 0.8 & 0.9 & 84.3 & \textbf{61.6} \\
\midrule

\multirow{3}{*}{Meta-Llama-3-8B-Instruct}
& Trivial name & \textbf{31.4} & \textbf{38.6} & 67.7 & 29.0 \\
& IUPAC name   & 3.1 & 6.6 & 89.4 & \textbf{57.4} \\
& SMILES       & 0.6 & 0.9 & \textbf{89.7} & 47.0 \\
\midrule

\multirow{3}{*}{Llama-3.1-8B-Instruct}
& Trivial name & \textbf{60.9} & \textbf{39.1} & 65.5 & 25.7 \\
& IUPAC name   & 2.8 & 6.0 & \textbf{90.5} & \textbf{49.9} \\
& SMILES       & 0.7 & 1.2 & 90.4 & 38.4 \\
\bottomrule
\end{tabular}
}
\end{table*}

\paragraph{Additional structure-computable tasks.}
The benchmark additionally includes molecular-formula prediction, total ring
count, and heavy-atom count. These tasks do not exhibit the stable identifier
ordering observed for benzene detection and nitrogen counting (Table~\ref{tab:additional_structure_tasks}). Formula
prediction strongly favors IUPAC and SMILES for ChemDFM-v2.0-14B, but the
trivial name performs best for most other models. Ring-count performance
similarly varies in both magnitude and preferred identifier, while heavy-atom
accuracy is generally low. We therefore report these tasks for completeness.

\begin{table*}[t]
\centering
\small
\caption{
\textbf{Additional structure-computable tasks.}
Accuracy (\%) for molecular-formula prediction, total ring count, and
heavy-atom count under each molecular identifier. Majority-class baselines
for ring count and heavy-atom count over the 896 molecules are 23.9\% and
5.5\%, respectively. Best performance within each model and task is shown in
bold.
}

\label{tab:additional_structure_tasks}
\resizebox{\textwidth}{!}{
\begin{tabular}{llccc}
\toprule
\textbf{Model} & \textbf{Identifier}
& \textbf{Formula}
& \textbf{Ring count}
& \textbf{Heavy atoms} \\
\midrule

\multirow{3}{*}{ChemDFM-v2.0-14B}
& Trivial name & 7.7 & \textbf{24.4} & 6.8 \\
& IUPAC name   & \textbf{56.8} & 16.9 & \textbf{10.6} \\
& SMILES       & 43.5 & 7.7 & 9.8 \\
\midrule

\multirow{3}{*}{ChemDFM-v1.5-8B}
& Trivial name & \textbf{0.0} & 23.0 & \textbf{3.1} \\
& IUPAC name   & \textbf{0.0} & 26.0 & 2.3 \\
& SMILES       & \textbf{0.0} & \textbf{48.1} & 1.5 \\
\midrule

\multirow{3}{*}{Qwen2.5-14B-Instruct}
& Trivial name & \textbf{12.7} & 24.3 & 2.7 \\
& IUPAC name   & 5.2 & 19.9 & \textbf{4.8} \\
& SMILES       & 1.1 & \textbf{39.3} & 3.3 \\
\midrule

\multirow{3}{*}{Qwen2.5-7B-Instruct}
& Trivial name & \textbf{7.0} & 26.8 & 0.9 \\
& IUPAC name   & 3.9 & \textbf{39.4} & \textbf{3.9} \\
& SMILES       & 0.2 & 14.3 & 1.2 \\
\midrule

\multirow{3}{*}{Qwen2-72B-Instruct}
& Trivial name & \textbf{16.7} & 32.5 & 4.0 \\
& IUPAC name   & 7.1 & \textbf{55.1} & 8.5 \\
& SMILES       & 2.9 & 48.4 & \textbf{10.9} \\
\midrule

\multirow{3}{*}{Meta-Llama-3-8B-Instruct}
& Trivial name & \textbf{8.7} & \textbf{24.9} & 2.3 \\
& IUPAC name   & 4.4 & 20.9 & 4.9 \\
& SMILES       & 1.1 & 18.3 & \textbf{7.5} \\
\midrule

\multirow{3}{*}{Llama-3.1-8B-Instruct}
& Trivial name & \textbf{10.2} & 27.0 & 0.4 \\
& IUPAC name   & 5.7 & \textbf{30.2} & 6.1 \\
& SMILES       & 2.5 & 22.8 & \textbf{7.8} \\
\bottomrule
\end{tabular}
}
\end{table*}

\paragraph{Paired significance testing.}
Because every molecule is evaluated under each identifier condition, the
observations are paired. We use McNemar's
test~\citep{mcnemar1947note} to determine whether the directional differences
between identifier conditions occur systematically across molecules.
For two identifier conditions \(A\) and \(B\), let \(n_{10}\) denote the
number of molecules answered correctly under \(A\) but incorrectly under
\(B\), and \(n_{01}\) the number answered incorrectly under \(A\) but
correctly under \(B\). Under the null hypothesis of no directional difference,
the two types of discordant pair are expected to occur equally often. We use
the continuity-corrected statistic:
\[
\chi^2 =
\frac{(|n_{10}-n_{01}|-1)^2}{n_{10}+n_{01}},
\]
which is compared with a \(\chi^2\) distribution with one degree of freedom.

Table~\ref{tab:mcnemar_identifier} reports the paired results for
identification, therapeutic indication, and benzene-ring detection. The two
comparisons emphasized in the main text, trivial name versus SMILES for
therapeutic indication and SMILES versus trivial name for benzene
detection, are significant for all seven models (\(p<10^{-7}\)). Thus, the
aggregate differences reflect directional asymmetries on the same paired
molecules rather than differences between evaluation sets.
\begin{table*}[t]
\centering
\small
\caption{
\textbf{Paired McNemar tests for cross-identifier differences.}
Each entry reports \(n_{10}\) versus \(n_{01}\), where \(n_{10}\) favors the
direction stated in the column heading, followed by the continuity-corrected
McNemar \(p\)-value.
}
\label{tab:mcnemar_identifier}
\resizebox{\textwidth}{!}{
\begin{tabular}{lccc}
\toprule
\textbf{Model}
& \textbf{Identify: Name \(>\) SMILES}
& \textbf{Indication: Name \(>\) SMILES}
& \textbf{Benzene: SMILES \(>\) Name} \\
\midrule

ChemDFM-v2.0-14B
& 775 vs.\ 1 \; (\(p=2\times10^{-169}\))
& 333 vs.\ 4 \; (\(p=2\times10^{-71}\))
& 198 vs.\ 37 \; (\(p=2\times10^{-25}\)) \\

ChemDFM-v1.5-8B
& 598 vs.\ 19 \; (\(p=9\times10^{-120}\))
& 183 vs.\ 14 \; (\(p=5\times10^{-33}\))
& 212 vs.\ 88 \; (\(p=1\times10^{-12}\)) \\

Qwen2.5-14B-Instruct
& 765 vs.\ 1 \; (\(p=3\times10^{-167}\))
& 331 vs.\ 3 \; (\(p=1\times10^{-71}\))
& 322 vs.\ 68 \; (\(p=1\times10^{-37}\)) \\

Qwen2.5-7B-Instruct
& 370 vs.\ 1 \; (\(p=2\times10^{-81}\))
& 299 vs.\ 4 \; (\(p=5\times10^{-64}\))
& 292 vs.\ 93 \; (\(p=6\times10^{-24}\)) \\

Qwen2-72B-Instruct
& 553 vs.\ 4 \; (\(p=2.9\times10^{-119}\))
& 384 vs.\ 2 \; (\(p=8.9\times10^{-84}\))
& 215 vs.\ 96 \; (\(p=2.2\times10^{-11}\)) \\

Meta-Llama-3-8B-Instruct
& 281 vs.\ 5 \; (\(p=2\times10^{-59}\))
& 338 vs.\ 0 \; (\(p=5\times10^{-75}\))
& 248 vs.\ 51 \; (\(p=9\times10^{-30}\)) \\

Llama-3.1-8B-Instruct
& 542 vs.\ 2 \; (\(p=4\times10^{-118}\))
& 343 vs.\ 4 \; (\(p=1\times10^{-73}\))
& 278 vs.\ 55 \; (\(p=5\times10^{-34}\)) \\
\bottomrule
\end{tabular}
}
\end{table*}
\subsection{Recognition and Cross-Identifier Binding}
\label{app:recognition}

This section provides the implementation details for the recognition controls
used in Section~\ref{sec:recognition}. We evaluate whether a model can recover
the trivial name associated with a SMILES string using two complementary
criteria: two-alternative forced choice (2AFC) and free-form generation.

\paragraph{2AFC recognition.}
For each molecule, the model is shown its canonical SMILES together with the
correct trivial name and one incorrect candidate:
\begin{quote}
\small
\texttt{A compound is given as SMILES \{SMILES\}.}\\
\texttt{Which name identifies it?}\\
\texttt{A) \{option\_a\}}\\
\texttt{B) \{option\_b\}}\\
\texttt{Reply with a single letter only.}
\end{quote}
The order of the correct and incorrect candidates is randomized. Responses are
scored by extracting a standalone \texttt{A} or \texttt{B}.

We construct three decoy conditions. In the \emph{near-analogue} condition,
the incorrect candidate is sampled from molecules with Morgan-fingerprint
similarity of at least 0.4. We use radius-2, 2,048-bit Morgan fingerprints
\citep{rogers2010extended} and compute molecular similarity using the
Tanimoto coefficient \citep{bajusz2015tanimoto}. If no molecule satisfies
the similarity threshold, the decoy is sampled from the five nearest
molecules. In the \emph{random} condition, the decoy is sampled uniformly
from all other eligible molecules. In the \emph{gross-mismatch} condition,
the decoy is sampled from molecules whose heavy-atom count differs from the
target by more than a factor of two, falling back to the full eligible pool
when necessary. Heavy-atom counts follow the RDKit-based definition in
Appendix~\ref{app:evaluation_protocol1}. Sampling uses a fixed random seed of 0.
Decoys are sampled independently across conditions and therefore need not be
distinct. A molecule is considered recognized under 2AFC only if the model
answers correctly under all three decoy conditions.

\paragraph{Free-form recognition.}
The stricter free-form control uses the standard SMILES-to-name identification
task defined in Appendix~\ref{app:evaluation_protocol1}: given only the
canonical SMILES, the model must generate the molecule's trivial name without
candidate answers. Predictions and gold names are normalized by lowercasing,
removing possessive suffixes and non-alphanumeric punctuation, normalizing
whitespace, and lightly singularizing words longer than four characters that
end in \texttt{s} but not \texttt{ss}. Recognition succeeds when the
normalized gold name occurs as a word-boundary-delimited substring of the
normalized response.

\paragraph{Recognized-subset evaluation.}
For either recognition criterion, let \(R\) denote the set of recognized
molecules that also appear in the therapeutic-indication evaluation.
Table~\ref{tab:recognition} reports three quantities. \textit{Name} is
therapeutic-indication accuracy when molecules in \(R\) are queried through
their trivial names; \textit{SMILES} is accuracy on the same molecules when
queried through canonical SMILES; and \textit{Rest} is canonical-SMILES
accuracy on molecules outside \(R\). Thus, the Name and SMILES columns always
compare exactly the same recognized molecules.
All therapeutic-indication results reported in Table~\ref{tab:recognition}
use the strict indication-scoring criterion defined in
Appendix~\ref{app:evaluation_protocol1}. 

\section{Identifier-Bridge and Conflict Experiments}
\label{app:bridge_conflict}

This section provides the full protocols and additional results supporting
Section~\ref{sec:unreachable}.

\subsection{Identifier-Bridge Protocol}
\label{app:bridge}

For each molecule with canonical SMILES \(s\) and trivial name \(n\), we
evaluate therapeutic-indication retrieval under three matched conditions.
The exact prompts are:

\paragraph{SMILES only.}
\begin{quote}
\small
\texttt{Consider the molecule with SMILES \{SMILES\}.}\\
\texttt{What medical condition is this substance primarily used to treat?}\\
\texttt{Answer in English. Reply with the condition only. Do not explain.}
\end{quote}

\paragraph{SMILES + name.}
\begin{quote}
\small
\texttt{Consider the molecule with SMILES \{SMILES\}, which is known as
"\{NAME\}".}\\
\texttt{What medical condition is this substance primarily used to treat?}\\
\texttt{Answer in English. Reply with the condition only. Do not explain.}
\end{quote}

\paragraph{Name only.}
\begin{quote}
\small
\texttt{Consider the molecule commonly known as "\{NAME\}".}\\
\texttt{What medical condition is this substance primarily used to treat?}\\
\texttt{Answer in English. Reply with the condition only. Do not explain.}
\end{quote}

The SMILES+name condition therefore supplies only the correspondence between
the structural and lexical identifiers; it does not reveal the therapeutic
indication. Each fact and identifier condition is queried independently.
Decoding and indication scoring follow Appendix~\ref{app:evaluation_protocol1}.

We quantify how much of the SMILES--name access gap is closed as
\[
\mathrm{Recovery}
=
\frac{A_{\text{SMILES+Name}}-A_{\text{SMILES}}}
     {A_{\text{Name}}-A_{\text{SMILES}}}
\times 100,
\]
where \(A\) denotes therapeutic-indication accuracy. A value of \(100\%\)
means that supplying the name closes the entire gap between SMILES-only and
name-only retrieval.

\paragraph{Reverse bridge.}
We also test the converse intervention for two structure-computable
properties: benzene-ring presence and nitrogen-atom count. Here we compare
the trivial name alone, the trivial name augmented with the canonical SMILES,
and the SMILES alone. For example, the benzene prompts use
\texttt{Consider the molecule commonly known as "\{NAME\}".},
\texttt{Consider the molecule commonly known as "\{NAME\}", whose SMILES is
\{SMILES\}.}, and
\texttt{Consider the molecule with SMILES \{SMILES\}.}, followed by the same
benzene-ring question used in the benchmark. Nitrogen counting uses the
corresponding three identifier prefixes followed by the standard
nitrogen-count question.

Table~\ref{tab:reverse_bridge} reports the complete saved reverse-bridge
results for Qwen2-72B-Instruct. Adding SMILES to the lexical prompt closes,
and slightly exceeds, the name--SMILES gap for both properties.

\begin{table}[t]
\centering
\caption{\textbf{Reverse identifier bridge.}
Accuracy (\%) for Qwen2-72B-Instruct. Reverse recovery is the fraction of the
name-to-SMILES accuracy gap closed by adding SMILES to the name. Values above
\(100\%\) indicate that the combined prompt slightly exceeds SMILES alone.}
\label{tab:reverse_bridge}
\small
\setlength{\tabcolsep}{5pt}
\begin{tabular}{lrrrr}
\toprule
Property & Name & Name + SMILES & SMILES & Recovery \\
\midrule
Benzene   & 71.1 & 85.7 & 84.3 & 111.0 \\
Nitrogens & 39.6 & 62.3 & 61.5 & 103.6 \\
\bottomrule
\end{tabular}
\end{table}

\subsection{Conflicting-Identifier Protocol}
\label{app:conflict_protocol}

To test whether lexical and structural identifiers are reconciled into a
single entity representation, we pair the canonical SMILES of molecule \(A\)
with the trivial name of a different molecule \(B\). For each molecule, the
decoy is sampled from molecules whose normalized reference therapeutic
indication differs from that of the true molecule. One candidate is drawn
using a fixed random seed of \(0\).

The conflicting identifier is presented as
\begin{quote}
\small
\texttt{the molecule with SMILES \{SMILES\}, which is known as
"\{DECOY\_NAME\}"}
\end{quote}
and each queried property is evaluated in a separate prompt. For identity,
therapeutic indication, benzene-ring presence, and nitrogen count, we append
the same task-specific question used in the corresponding benchmark
evaluation. Thus, the experiment does not ask multiple properties jointly.

For identity, we measure how often the response follows the true molecule
\(A\) specified by the SMILES versus the conflicting lexical identity \(B\).
For therapeutic indication, we measure retrieval of \(B\)'s indication and
compare it with retrieval when \(B\) is queried through its name alone. For
structure-computable properties, we compare accuracy under conflict with the
corresponding SMILES-only condition for \(A\).

Table~\ref{tab:conflict_full} reports the cross-model results used in
Section~\ref{sec:conflict}. Entity-associated outputs largely follow the
incorrect lexical identity, whereas benzene judgments remain close to the
SMILES-only condition.
\begin{table*}[t]
\centering
\caption{\textbf{Conflicting identifiers control different outputs:} Each input pairs the SMILES of molecule \(A\) with the incorrect trivial name
of molecule \(B\). Identity reports retrieval of \(A\) versus \(B\);
therapeutic indication compares retrieval of \(B\)'s indication under
conflict with retrieval from \(B\)'s name alone; benzene compares accuracy
under conflict with \(A\)'s SMILES-only condition. All values are percentages.}
\label{tab:conflict_full}
\small
\setlength{\tabcolsep}{4pt}
\begin{tabular}{lccc}
\toprule
Model
& Identity: \(A/B\)
& Indication of \(B\): conflict / name
& Benzene of \(A\): conflict / SMILES \\
\midrule
ChemDFM-v2.0-14B
& 1.0 / 85.5 & 32.3 / 38.6 & 95.3 / 94.5 \\
ChemDFM-v1.5-8B
& 0.6 / 73.4 & 17.4 / 23.0 & 80.5 / 85.0 \\
Qwen2.5-14B-Instruct
& 0.0 / 81.1 & 35.9 / 37.8 & 84.9 / 86.9 \\
Qwen2.5-7B-Instruct
& 0.0 / 60.4 & 32.1 / 33.9 & 75.9 / 80.2 \\
Llama-3.1-8B-Instruct
& 0.0 / 86.8 & 37.3 / 39.0 & 89.3 / 90.3 \\
Qwen2-72B-Instruct
& 0.0 / 77.8 & 43.2 / 44.1 & 83.9 / 84.3 \\
\bottomrule
\end{tabular}
\end{table*}
For Qwen2-72B-Instruct, the same conflicting prompts additionally retain
59.9\% nitrogen-count accuracy, while the supplied decoy name is recovered
in 77.8\% of identity queries and the true name in 0.0\%. This provides a
second structure-computable example of the same split between lexical and
structural access.

\subsection{Direct Detection of Identifier Conflicts}
\label{app:conflict_detection}

We separately test whether models can explicitly recognize that a supplied
SMILES and trivial name are inconsistent. Each canonical SMILES is evaluated
with its correct name and with three types of incorrect names.

\paragraph{Mismatch construction.}
For the near-analogue condition, we represent each molecule using a radius-2,
2,048-bit Morgan fingerprint~\citep{rogers2010extended}, a circular molecular
fingerprint that encodes local structural subpatterns, and measure pairwise
similarity using the Tanimoto coefficient~\citep{bajusz2015tanimoto}. The
incorrect name is sampled uniformly from molecules with Tanimoto similarity at
least 0.4 to the target. If no such candidate exists, the decoy is sampled from
the five nearest molecules, with all other eligible molecules used as a final
fallback.

For the random condition, the decoy is sampled uniformly from all other
molecules with valid fingerprints. For the gross-mismatch condition, the decoy
is sampled from molecules whose heavy-atom count differs from the target by
more than a factor of two, falling back to the full eligible pool when
necessary. Sampling uses a fixed random seed of 0; decoys are sampled
independently across conditions and therefore need not be distinct.

\paragraph{Direct consistency check.}
The exact prompt is:
\begin{quote}
\texttt{A compound is given as SMILES \{SMILES\} and is labelled}\\
\texttt{"\{LABEL\}".}\\
\texttt{Does this label correctly identify the structure shown?}\\
\texttt{Reply with exactly one word: yes or no.}
\end{quote}

The response parser evaluates negative cues before positive cues. Negative cues
include \texttt{no}, \texttt{not}, \texttt{incorrect}, \texttt{mismatch},
\texttt{wrong}, \texttt{different}, and \texttt{false}; positive cues include
\texttt{yes}, \texttt{correct}, \texttt{match}, and \texttt{true}.

\paragraph{Near-analogue 2AFC.}
We additionally ask the model to choose between the true trivial name and the
selected near-analogue:
\begin{quote}
\texttt{A compound is given as SMILES \{SMILES\}.}\\
\texttt{Which name identifies it?}\\
\texttt{A) \{OPTION A\}}\\
\texttt{B) \{OPTION B\}}\\
\texttt{Reply with a single letter only.}
\end{quote}
The order of the true and incorrect candidates is randomized.

\paragraph{Discrimination metric.}
Mismatch-rejection rate alone can be misleading because a model could obtain a
high rejection rate simply by rejecting most name--SMILES pairs. We therefore
also report the signal-detection sensitivity index \(d'\)~\citep{stanislaw1999calculation},
which accounts for false alarms on correctly matched pairs. Let \(H\) denote
the probability of rejecting a mismatched pair and \(F\) the probability of
incorrectly rejecting a correctly matched pair. We compute
\[
d' = Z(H) - Z(F),
\]
where \(Z\) is the inverse standard-normal cumulative distribution function.
Larger \(d'\) indicates better discrimination between matched and mismatched
identifier pairs. Before applying \(Z\), both probabilities are clipped to
\([0.001, 0.999]\). No sample-size-dependent correction is applied.

\begin{table}[t]
\centering
\caption{\textbf{Direct detection of identifier conflicts.}
``Correct accepted'' is the percentage of valid name--SMILES pairs that the
model accepts as matching. ``Near reject'' and ``Gross reject'' are the
percentages of near-analogue and gross mismatches correctly rejected,
respectively. The corresponding \(d'\) values measure discrimination while
accounting for false alarms on correctly matched pairs. ``Near 2AFC'' is
accuracy when choosing between the true trivial name and a structurally similar
incorrect name. All rates are percentages except \(d'\).}
\label{tab:conflict_detection}
\small
\setlength{\tabcolsep}{3.5pt}
\begin{tabular}{lrrrrrr}
\toprule
Model
& Correct
& Near
& Near
& Gross
& Gross
& Near \\
& accepted
& reject
& \(d'\)
& reject
& \(d'\)
& 2AFC \\
\midrule
ChemDFM-v2.0-14B
& 93.6 & 36.9 & 1.19 & 84.3 & 2.53 & 73.9 \\

ChemDFM-v1.5-8B
& 65.5 & 43.1 & 0.22 & 62.0 & 0.70 & 71.9 \\

Qwen2.5-14B-Instruct
& 76.5 & 38.6 & 0.43 & 59.7 & 0.97 & 62.8 \\

Qwen2.5-7B-Instruct
& 87.3 & 17.5 & 0.21 & 28.5 & 0.57 & 56.4 \\

Qwen2-72B-Instruct
& 91.7 & 22.2 & 0.62 & 47.0 & 1.31 & 65.5 \\
\bottomrule
\end{tabular}
\end{table}

Gross mismatches are substantially easier to detect than structurally similar
ones. For ChemDFM-v2.0-14B, rejection falls from 84.3\% for gross mismatches
to 36.9\% for near analogues, with \(d'\) decreasing from 2.53 to 1.19.
Across the remaining models, near-analogue rejection ranges from 17.5\% to
43.1\%, with \(d'\) between 0.21 and 0.62. Near-analogue 2AFC accuracy is only
56.4--73.9\%, despite a 50\% chance baseline.

For Qwen2-72B-Instruct, for which complete saved random-mismatch
discrimination results are also available, random mismatches fall between the
two extremes: rejection is 32.6\% with \(d' = 0.94\), compared with 22.2\%
and 0.62 for near analogues and 47.0\% and 1.31 for gross mismatches.
Random-mismatch 2AFC accuracy is 70.6\%.

\section{Activation-Patching Details}
\label{app:patching}

This section provides the implementation details for the causal intervention
in Section~\ref{sec:patching}.

\subsection{Source and Target Construction}

For each molecule, we construct matched prompts querying the same therapeutic
indication through either the trivial name (\emph{source}) or canonical
SMILES (\emph{target}):

\begin{quote}
\small
\texttt{The molecule commonly known as "\{NAME\}" is primarily used to treat}
\end{quote}

\begin{quote}
\small
\texttt{The molecule with SMILES \{SMILES\} is primarily used to treat}
\end{quote}

The gold continuation is the reference indication preceded by a single
space. For chat models, the source or target string is placed in the user
turn with the system message \texttt{You are a helpful assistant.} and the
model's chat template is applied with a generation prompt. The gold
continuation is then scored autoregressively under teacher forcing.

\subsection{Item Selection}

We first compute the unpatched mean gold-token log-probability under the
trivial-name and SMILES prompts. We retain only molecules for which lexical
access is measurably stronger,

\[
\mathrm{lp}_{\mathrm{name}}
-
\mathrm{lp}_{\mathrm{SMILES}}
\ge 0.20
\]

nats per gold token. This ensures that recovery is evaluated on examples that
exhibit a pre-existing name--SMILES access gap. For Qwen2-72B-Instruct,
491 of the 896 benchmark molecules satisfy this criterion, spanning 176
distinct indication strings.

\subsection{Activation Intervention}

At a tested transformer block \(\ell\), we capture the block-output residual
state at the final prompt token in the trivial-name source run. We then run
the matched SMILES prompt and replace the complete block-output vector at
its final prompt position with the captured source vector:

\[
h^{(\ell)}_{\mathrm{SMILES}}
\leftarrow
h^{(\ell)}_{\mathrm{name}}.
\]

All other token positions remain unchanged, and computation proceeds normally
through the remaining transformer blocks. Each layer intervention is
evaluated independently.

Layers are searched using a coarse-to-fine procedure: an initial regularly
spaced sweep identifies a coarse peak, after which the surrounding layers are
evaluated more densely. The reported peak is the tested layer with the
largest mean recovery. Relative depths in the main text normalize the selected
layer index by model depth.

\subsection{Teacher-Forced Scoring and Recovery}

For a gold continuation \(g_1,\ldots,g_T\), we compute mean token
log-probability

\[
\mathrm{lp}
=
\frac{1}{T}
\sum_{t=1}^{T}
\log p(g_t \mid x,g_{<t}),
\]

where \(x\) is the rendered prompt. Gold tokens are scored autoregressively
under teacher forcing. Logits are converted to float32 before applying
\texttt{log\_softmax}.

Recovery is computed separately for every molecule \(i\) and tested layer
\(\ell\):

\[
R_{i,\ell}
=
\frac{
\mathrm{lp}^{\mathrm{patched}}_{i,\ell}
-
\mathrm{lp}^{\mathrm{SMILES}}_i
}{
\max(
\mathrm{lp}^{\mathrm{name}}_i
-
\mathrm{lp}^{\mathrm{SMILES}}_i,
10^{-6})
}.
\]

The reported layer-level recovery is the arithmetic mean of
\(R_{i,\ell}\) across selected molecules, multiplied by \(100\).
Thus, \(100\%\) indicates that patching closes the original name--SMILES
gap on average, while values above \(100\%\) indicate that the patched
computation exceeds the corresponding name-conditioned score. Standard
errors are computed across the per-molecule recovery values.

\subsection{Different-Entity Control}

To test whether recovery arises from injecting arbitrary lexical activity,
each focal molecule is assigned a control molecule with a different reference
indication. The control is sampled uniformly from the selected items using a
fixed random seed of \(0\). The assignment is made once and reused at every
tested layer.

At each layer, we therefore repeat the intervention using the control
molecule's trivial-name activation while retaining the focal molecule's
SMILES target and gold indication. Control recovery is computed using the
same denominator as the matched intervention:

\[
R^{\mathrm{ctrl}}_{i,\ell}
=
\frac{
\mathrm{lp}^{\mathrm{ctrl}}_{i,\ell}
-
\mathrm{lp}^{\mathrm{SMILES}}_i
}{
\max(
\mathrm{lp}^{\mathrm{name}}_i
-
\mathrm{lp}^{\mathrm{SMILES}}_i,
10^{-6})
}.
\]

The consistently negative controls in
Table~\ref{tab:activation_patching} show that recovery is not produced merely
by inserting an arbitrary lexical-state vector.

\paragraph{Qwen2-72B-Instruct layer sweep.}
For the 80-layer Qwen2-72B-Instruct model, the initial sweep evaluates every
fourth layer. After a coarse peak at layer 68, the surrounding region is
searched densely, yielding 32 tested layers in total. The final peak occurs
at layer 66, with mean recovery \(138.4\%\pm8.0\) percentage points (SEM);
the corresponding different-entity control is \(-30.9\%\).

\section{Cross-Identifier Representation Analysis}
\label{app:representation}

This section provides the implementation details and complete results for the
representation analysis in Section~\ref{sec:representations}. 

\subsection{Representation Extraction}
\label{app:representation_extraction}

For each of the 896 benchmark molecules, we construct three identifier-specific
prompts:
\begin{quote}
\small
\texttt{the molecule commonly known as "\{NAME\}"}

\texttt{the molecule named \{IUPAC\_NAME\}}

\texttt{the molecule with SMILES \{SMILES\}}
\end{quote}

Each complete prompt is tokenized using the checkpoint's tokenizer with its
default special-token behavior. Batches are right-padded and truncated to at
most 512 tokens. Hidden states are extracted for every state returned by the
model.

For molecule \(i\), identifier form \(a\), and hidden-state index \(\ell\),
let \(\mathbf H^{(\ell)}_{i,a,t}\) denote the hidden state at token position
\(t\), and let \(m_{i,a,t}\in\{0,1\}\) be the corresponding attention-mask
value. We represent the prompt by the arithmetic mean over all non-padding
tokens:
\[
\mathbf h^{(\ell)}_{i,a}
=
\frac{
    \sum_t m_{i,a,t}\mathbf H^{(\ell)}_{i,a,t}
}{
    \sum_t m_{i,a,t}
}.
\]

Thus, the representation is a mean over the complete non-padding prompt rather
than over the identifier tokens alone. It includes the fixed prompt prefix,
the molecular identifier, punctuation such as quotation marks when present,
and any special tokens added by the tokenizer. Padding tokens are excluded.

We use the hidden-state sequence returned directly by the model. Hidden-state
index \(0\) corresponds to the embedding output, followed by the states
returned after successive transformer computations. We therefore report the
exact hidden-state index used by the implementation rather than converting it
to a separately defined block number.

The extraction pipeline also supports pooling at the final non-padding prompt
token, but all results reported in Section~\ref{sec:representations},
Figure~\ref{fig:representation}, and this appendix use the mean-pooled
representations defined above. We do not select the pooling strategy
separately for different identifier pairs.

\subsection{Cross-Identifier Retrieval}
\label{app:representation_retrieval}

We evaluate whether the representation of a molecule under one identifier can
retrieve the representation of the same molecule under another identifier.
The candidate set contains \(N=896\) molecule rows.

Before retrieval, representations are centered separately for each identifier
form and hidden-state index. For identifier form \(a\),
\[
\boldsymbol{\mu}^{(\ell)}_a
=
\frac{1}{N}
\sum_{i=1}^{N}
\mathbf h^{(\ell)}_{i,a},
\]
and each centered representation is L2-normalized:
\[
\tilde{\mathbf h}^{(\ell)}_{i,a}
=
\frac{
    \mathbf h^{(\ell)}_{i,a}
    -
    \boldsymbol{\mu}^{(\ell)}_a
}{
    \left\|
    \mathbf h^{(\ell)}_{i,a}
    -
    \boldsymbol{\mu}^{(\ell)}_a
    \right\|_2
    + 10^{-8}
}.
\]

For a query molecule \(i\) represented under identifier \(a\), retrieval under
identifier \(b\) is
\[
\hat{j}
=
\arg\max_{j\in\{1,\ldots,N\}}
\left(
\tilde{\mathbf h}^{(\ell)}_{i,a}
\right)^{\!\top}
\tilde{\mathbf h}^{(\ell)}_{j,b}.
\]

Because the vectors are L2-normalized, this dot product is cosine similarity
after mean-centering. No PCA, whitening, or other representation
transformation is applied.

Top-1 retrieval is correct when \(\hat{j}=i\). With 896 candidate rows,
uniform chance performance is
\[
\frac{1}{896} \approx 0.11\%.
\]

Throughout the paper, \textbf{Name--SMILES} denotes the directional retrieval
task
\[
\text{trivial name} \rightarrow \text{SMILES},
\]
and \textbf{IUPAC--SMILES} denotes
\[
\text{IUPAC name} \rightarrow \text{SMILES}.
\]
The reported metric is not an average over the two retrieval directions.

The 896-row candidate set contains distinct molecule identifiers, trivial
names, and canonical SMILES. Two IUPAC strings each occur for two distinct
molecule rows, yielding 894 distinct IUPAC strings overall. Evaluation remains
row-index based: the representation corresponding to the same molecule row is
treated as the gold target.

\subsection{Layer-Wise Peak Results}
\label{app:representation_results}

Top-1 retrieval is computed independently at every saved hidden-state index.
For each model and identifier pair, we report the maximum retrieval accuracy
over these indices. Different identifier pairs may therefore attain their
maximum at different depths.

Table~\ref{tab:representation_full} reports the complete mean-pooled,
forward-retrieval results together with the hidden-state index at which each
maximum occurs.

\begin{table}[t]
\centering
\caption{\textbf{Peak cross-identifier representation retrieval.}
Top-1 retrieval accuracy (\%) among 896 molecule rows using mean-pooled
whole-prompt hidden representations. Name--SMILES denotes trivial
name \(\rightarrow\) SMILES retrieval, and IUPAC--SMILES denotes
IUPAC name \(\rightarrow\) SMILES retrieval. ``Peak idx.'' is the
hidden-state index attaining the maximum accuracy; index \(0\) is the
embedding output. ``States'' is the total number of returned hidden states.
Chance performance is \(0.11\%\).}
\label{tab:representation_full}
\small
\setlength{\tabcolsep}{3.5pt}
\begin{tabular}{lrrrrr}
\toprule
& \multicolumn{2}{c}{Name \(\rightarrow\) SMILES}
& \multicolumn{2}{c}{IUPAC \(\rightarrow\) SMILES}
& \\
\cmidrule(lr){2-3}
\cmidrule(lr){4-5}
Model
& Top-1
& Peak idx.
& Top-1
& Peak idx.
& States \\
\midrule
ChemDFM-v2.0-14B
& 0.8 & 28
& 13.5 & 31
& 49 \\

ChemDFM-v1.5-8B
& 2.8 & 24
& 15.0 & 32
& 33 \\

Qwen2.5-14B-Instruct
& 0.7 & 36
& 2.6 & 30
& 49 \\

Qwen2.5-7B-Instruct
& 0.7 & 3
& 1.8 & 25
& 29 \\

Meta-Llama-3-8B-Instruct
& 0.8 & 13
& 3.9 & 18
& 33 \\

Llama-3.1-8B-Instruct
& 0.9 & 6
& 4.0 & 18
& 33 \\
\bottomrule
\end{tabular}
\end{table}

\section{Lookup-Based Lexical Routing: Evaluation and Controls}
\label{app:resolver}

This section provides the evaluation setup, scoring procedure, and
self-resolution control for the lexical-routing experiment in
Section~\ref{sec:lookup}.

\subsection{Multi-Relation Evaluation Set}
\label{app:resolver_set}

We evaluate four entity-associated relations: therapeutic indication,
molecular target, mechanism of action (MoA), and ATC therapeutic class.
Starting from the cleaned benchmark collection described in
Appendix~\ref{app:benchmark_details}, we augment each molecule with available
ChEMBL relation annotations and retain molecules for which at least two of
the four relations are available. This yields 891 molecules and 2,955
molecule--relation pairs.

\begin{table}[t]
\centering
\caption{\textbf{Coverage of the multi-relation evaluation set.}
Number of molecules with at least one accepted answer for each relation.}
\label{tab:resolution_relation_coverage}
\small
\begin{tabular}{lr}
\toprule
Relation & Molecules \\
\midrule
Therapeutic indication & 891 \\
Molecular target        & 572 \\
Mechanism of action     & 651 \\
ATC therapeutic class   & 841 \\
\midrule
Total molecule--relation pairs & 2,955 \\
\bottomrule
\end{tabular}
\end{table}

Therapeutic indications are obtained from ChEMBL
\texttt{drug\_indication} records with
\texttt{max\_phase\_for\_ind >= 4}. Molecular targets are obtained through
\texttt{mechanism.target\_chembl\_id} and mapped to the corresponding target
preferred name. Mechanism-of-action labels use
\texttt{mechanism\_of\_action}. For ATC, we consider up to the first three
available classifications and use the level-4 description, falling back to
the level-3 description when necessary. When multiple targets, mechanisms,
or ATC labels are available, all are retained as acceptable answers.

\subsection{Lexical-Routing Conditions}
\label{app:resolver_conditions}

For each available molecule--relation pair, we evaluate four conditions:

\begin{enumerate}
    \item \textbf{SMILES}: query the relation directly from the molecule's
    canonical SMILES.

    \item \textbf{Self-resolution}: first ask the model to generate the
    molecule's common name from SMILES, then insert the generated name into
    the factual query.

    \item \textbf{Lookup}: supply the ChEMBL preferred name associated with
    the molecule together with its SMILES before asking the factual question.

    \item \textbf{Name}: query the relation directly from the ChEMBL
    preferred name.
\end{enumerate}

The lookup condition implements the routing path

\[
\mathrm{SMILES}
\;\xrightarrow{\text{name lookup}}\;
\mathrm{Lexical\ identity}
\;\xrightarrow{\text{LM}}\;
\mathrm{Fact}.
\]

The lookup supplies only the correspondence between the SMILES and the
lexical identifier; it contains no information about the queried relation.
In the implementation, the name is the ChEMBL preferred name already
associated with the molecule in the benchmark record. No live resolver API
is called during inference.

All factual queries use

\begin{quote}
\small
\texttt{Answer the question about the compound below.}\\
\texttt{Compound: \{REFERENCE\}}\\
\texttt{Question: \{QUESTION\}}\\
\texttt{Answer in English. Reply with the answer only. Do not explain.}
\end{quote}

The relation-specific questions are

\begin{quote}
\small
\texttt{what medical condition is it primarily used to treat?}\\
\texttt{what is its primary molecular target?}\\
\texttt{what is its mechanism of action?}\\
\texttt{what therapeutic class does it belong to?}
\end{quote}

with the following reference strings:

\begin{quote}
\small
\textbf{SMILES:}
\texttt{SMILES \{SMILES\}}\\[2pt]
\textbf{Self-resolution:}
\texttt{SMILES \{SMILES\} (identified as \{GENERATED\_NAME\})}\\[2pt]
\textbf{Lookup:}
\texttt{SMILES \{SMILES\} (resolved by name lookup to \{NAME\})}\\[2pt]
\textbf{Name:}
\texttt{\{NAME\}}
\end{quote}

For self-resolution, the model first receives

\begin{quote}
\small
\texttt{A database lookup returned this compound as SMILES: \{SMILES\}}\\
\texttt{What is its common generic name?}\\
\texttt{Answer in English. Reply with the name only.}
\end{quote}

Prompts are wrapped in the model's chat template when available, using the
system message \texttt{You are a helpful assistant.} Generation is greedy
(\texttt{do\_sample=False}) with at most 32 new tokens. Across 2,955
available molecule--relation pairs, the four factual-query conditions produce
11,820 factual queries.

\subsection{Answer Normalization and Scoring}
\label{app:resolver_scoring}

Predictions and accepted answers are normalized by lowercasing, removing
possessive suffixes, replacing non-alphanumeric punctuation with spaces,
collapsing whitespace, and lightly singularizing words longer than four
characters that end in \texttt{s} but not \texttt{ss}. Empty responses,
refusals, and responses longer than 14 normalized words are counted as
incorrect.

For all four relations, a response is correct when an accepted normalized
label occurs as a word-boundary-delimited phrase in the normalized
prediction. For molecular target and mechanism of action, we additionally
accept a prediction when it covers at least 60\% of the content words of an
accepted label after removing stopwords and tokens of length at most two.
Therapeutic indication and ATC class use only the phrase-matching criterion.

\subsection{Gap-Recovery Metric}
\label{app:resolver_recovery}

The main-text table reports aggregate factual accuracy under the SMILES,
lookup, and name conditions. To quantify how much of the SMILES--name access
gap is recovered by lexical routing, we compute

\[
\mathrm{GapRecovery}
=
100\times
\frac{
A_{\mathrm{lookup}}-A_{\mathrm{SMILES}}
}{
\max\!\left(
A_{\mathrm{name}}-A_{\mathrm{SMILES}},
10^{-9}
\right)
},
\]

where \(A_{\mathrm{SMILES}}\), \(A_{\mathrm{lookup}}\), and
\(A_{\mathrm{name}}\) denote aggregate accuracy under the corresponding
conditions. The complete aggregate results are reported in
Table~\ref{tab:lookup} in the main text.

\subsection{Self-Resolution Control}
\label{app:self_resolution}

Self-resolution asks whether the model can construct the missing lexical
bridge without an external identifier lookup. The model first predicts a name
from SMILES, and that generated name is then used in the downstream factual
query regardless of whether it matches the benchmark name.

\begin{table*}[t]
\centering
\caption{\textbf{Self-resolution results.}
SMILES-to-name identification accuracy and downstream factual accuracy (\%)
when the generated name is inserted into the query. Relation columns are
evaluated over molecules for which the corresponding relation is available.}
\label{tab:self_resolution_full}
\small
\setlength{\tabcolsep}{5pt}
\begin{tabular}{lrrrrr}
\toprule
Model &
Self-ID &
Indication &
Target &
MoA &
ATC \\
\midrule
ChemDFM-v2.0-14B
& 7.1 & 4.9 & 11.4 & 4.5 & 1.4 \\
ChemDFM-v1.5-8B
& 12.0 & 6.6 & 11.4 & 2.0 & 2.5 \\
Qwen2.5-14B-Instruct
& 0.3 & 1.2 & 4.0 & 3.1 & 1.1 \\
Qwen2.5-7B-Instruct
& 0.1 & 1.6 & 0.7 & 0.8 & 0.0 \\
Meta-Llama-3-8B-Instruct
& 0.7 & 2.6 & 3.3 & 1.1 & 1.8 \\
Llama-3.1-8B-Instruct
& 0.6 & 2.5 & 5.8 & 0.3 & 1.2 \\
\bottomrule
\end{tabular}
\end{table*}

SMILES-to-name identification ranges from \(0.1\%\) to \(12.0\%\) across the
six models. The low identification rates explain why self-resolution does not
provide the same access gains as supplying the correct lexical correspondence
directly.

\section{Knowledge Editing: Setup and Cross-Identifier Transfer}
\label{app:editing}

This section provides the experimental details for
Section~\ref{sec:editing}. We evaluate ROME and fine-tuning (FT) on
ChemDFM-v2.0-14B, Qwen2.5-14B-Instruct, Qwen2.5-7B-Instruct, and
Llama-3.1-8B-Instruct. Each counterfactual edit is introduced through the
molecule's trivial name and evaluated through the name, IUPAC name, and
SMILES.

\subsection{Edit Construction and Evaluation}

For each model, we retain molecules whose therapeutic indication is already
retrievable through the trivial name, yielding 389, 390, 351, and 389
molecules for ChemDFM-v2.0-14B, Qwen2.5-14B-Instruct,
Qwen2.5-7B-Instruct, and Llama-3.1-8B-Instruct, respectively.

For each molecule \(x\), we replace its original indication with a
counterfactual indication \(y^\ast\) sampled from another eligible molecule.
Candidates sharing normalized content words with any accepted true indication
of \(x\) are excluded before uniform sampling with seed \(0\). The original
raw indication string is used as the editing target. The edit request is

\begin{quote}
\texttt{\{common name\} is primarily used to treat}
\end{quote}

with \(y^\ast\) as the desired continuation. Each edit is applied and
evaluated independently, after which the modified weights are restored.

We evaluate the edited target under the following prompts:

\begin{table*}[t]
\centering
\caption{\textbf{Knowledge-editing evaluation prompts.}}
\label{tab:editing_prompts}
\small
\begin{tabular}{lp{0.72\textwidth}}
\toprule
Condition & Prompt \\
\midrule
Efficacy &
\texttt{\{Name.capitalize()\} is primarily used to treat} \\

IUPAC &
\texttt{The molecule named \{iupac\} is primarily used to treat} \\

SMILES &
\texttt{The molecule with SMILES \{smiles\} is primarily used to treat} \\

SMILES+Name &
\texttt{The molecule with SMILES \{smiles\}, which is known as
"\{name\}", is primarily used to treat} \\

Locality &
\texttt{\{Unrelated\_name.capitalize()\} is primarily used to treat} \\
\bottomrule
\end{tabular}
\end{table*}

Generation is deterministic with \texttt{do\_sample=False} and
\texttt{max\_new\_tokens=24}. We score the generated continuation using the
same normalized phrase-matching procedure as the main factual-retrieval
evaluation.

\paragraph{Editing implementation details.}
For each model and editing method, we first evaluated candidate editing
layers on a 30-example pilot sweep and used the pilot results to select the
layer for the full experiment. Because the pilot exhibited different
efficacy--locality trade-offs across models and methods, we did not impose
a single scalar selection metric across all settings. The resulting
ROME/FT layers were 24/24 for ChemDFM-v2.0-14B, 30/36 for
Qwen2.5-14B-Instruct, 17/14 for Qwen2.5-7B-Instruct, and 20/16 for
Llama-3.1-8B-Instruct, respectively. ROME edits the MLP down-projection at the selected layer and optimizes the
target vector for at most 25 steps using Adam with learning rate $0.5$,
weight decay $10^{-3}$.
FT optimizes the same down-projection for 25 steps using Adam with learning
rate $5\times10^{-4}$, batch size 1, zero weight decay, and element-wise
updates constrained to $5\times10^{-4}$ from the original weights.
Both methods use FP32 models, and evaluation uses greedy decoding with at
most 24 new tokens.

\subsection{Leakage-Corrected Cross-Identifier Transfer}

Retrieval of \(y^\ast\) through SMILES does not necessarily imply that the
edit transferred to the corresponding molecule: an edit may increase
production of \(y^\ast\) for unrelated molecules as well. We therefore
estimate this background effect using locality probes spanning molecular
similarities based on radius-2, 2,048-bit Morgan fingerprints and Tanimoto
similarity.

We define

\[
T_{\mathrm{leak}}
=
\operatorname{mean}
\bigl(
\texttt{post\_new}
\mid
\texttt{axis=locality}
\bigr),
\]

and report entity-specific SMILES transfer as

\[
T_{\mathrm{specific}}
=
T_{\mathrm{SMILES}}
-
T_{\mathrm{leak}}.
\]

Confidence intervals are obtained from 10,000 percentile-bootstrap resamples
with seed \texttt{20260921}, resampling at the edited-molecule level and
carrying the associated locality probes with each sampled molecule.

\subsection{Full Results}

\begin{table*}[t]
\centering
\caption{\textbf{Full cross-identifier knowledge-editing results.}
Leak is the pooled locality rate and Specific is SMILES minus Leak.
Brackets give 95\% bootstrap confidence intervals. Recovery is
SMILES+Name performance relative to name-conditioned efficacy.}
\label{tab:cross_identifier_editing_full}
\small
\setlength{\tabcolsep}{3pt}
\resizebox{\textwidth}{!}{
\begin{tabular}{llrrrrrrrr}
\toprule
Model & Method & \(n\) & Efficacy & IUPAC & SMILES & Leak &
Specific [95\% CI] & SMILES+Name & Recovery \\
\midrule
ChemDFM-v2.0-14B
& ROME & 389 & 52.2 & 24.2 & 22.1 & 19.8 &
+2.3 [$-0.9$,5.5] & 45.5 & 87.2 \\
& FT & 389 & 26.0 & 2.8 & 1.8 & 2.1 &
-0.3 [$-1.6$,1.2] & 25.2 & 97.0 \\

Qwen2.5-14B-Instruct
& ROME & 390 & 17.9 & 3.8 & 4.4 & 4.9 &
-0.5 [$-2.4$,1.4] & 12.3 & 68.6 \\
& FT & 390 & 29.0 & 14.4 & 11.5 & 8.0 &
+3.5 [0.9,6.3] & 26.7 & 92.0 \\

Qwen2.5-7B-Instruct
& ROME & 351 & 47.9 & 4.3 & 4.0 & 7.4 &
-3.4 [$-5.7$,$-0.9$] & 11.4 & 23.8 \\
& FT & 351 & 34.5 & 8.5 & 5.4 & 7.3 &
-1.9 [$-4.4$,0.8] & 21.7 & 62.8 \\

Llama-3.1-8B-Instruct
& ROME & 389 & 44.2 & 21.9 & 21.1 & 8.1 &
+13.0 [9.2,16.9] & 40.4 & 91.3 \\
& FT & 389 & 38.3 & 18.8 & 12.6 & 11.4 &
+1.2 [$-1.6$,4.1] & 32.1 & 83.9 \\
\bottomrule
\end{tabular}
}
\end{table*}

Across the eight configurations, six show no reliable positive
entity-specific transfer to SMILES beyond the leakage baseline. The two
exceptions are Qwen2.5-14B-Instruct with FT and Llama-3.1-8B-Instruct with
ROME, showing that standard editing can transfer across identifiers, but does
so inconsistently across models and methods.

\subsection{Conditioning on Successful Name Edits}

To rule out the possibility that weak transfer is caused simply by unsuccessful
edits, we repeat the analysis using only molecules for which the
counterfactual target is successfully retrieved through the trivial-name
efficacy prompt. Leakage is recomputed using locality probes associated only
with these successful edits.

\begin{table*}[t]
\centering
\caption{\textbf{Transfer conditioned on successful name edits.}
Specific is conditional SMILES transfer minus conditional leakage. Brackets
give 95\% bootstrap confidence intervals.}
\label{tab:editing_conditional}
\small
\setlength{\tabcolsep}{3pt}
\resizebox{\textwidth}{!}{
\begin{tabular}{llrcccc}
\toprule
Model & Method & \(n_{\mathrm{succ}}\) &
SMILES & Leak & Specific [95\% CI] & SMILES+Name \\
\midrule
ChemDFM-v2.0-14B
& ROME & 203 & 33.50 & 32.04 &
+1.46 [$-3.63$,6.51] & 67.98 \\
& FT & 101 & 4.95 & 5.94 &
-0.99 [$-5.12$,3.95] & 64.36 \\

Qwen2.5-14B-Instruct
& ROME & 70 & 18.57 & 20.99 &
-2.42 [$-10.98$,6.83] & 57.14 \\
& FT & 113 & 30.97 & 20.14 &
+10.83 [3.39,18.63] & 67.26 \\

Qwen2.5-7B-Instruct
& ROME & 168 & 7.14 & 13.23 &
-6.09 [$-10.73$,$-1.36$] & 22.62 \\
& FT & 121 & 12.40 & 16.99 &
-4.59 [$-10.96$,2.12] & 46.28 \\

Llama-3.1-8B-Instruct
& ROME & 172 & 38.37 & 11.95 &
+26.43 [19.63,33.19] & 74.42 \\
& FT & 149 & 26.85 & 22.74 &
+4.10 [$-1.77$,10.31] & 64.43 \\
\bottomrule
\end{tabular}
}
\end{table*}

The conclusion is unchanged after conditioning on successful name edits:
the same two configurations show clear positive specific transfer beyond
leakage, while the remaining six do not. Weak cross-identifier transfer
therefore cannot be explained simply by failure to install the edit through
the trivial name.

\section{Binding Self-Distillation: Experimental Details and Additional Results}
\label{app:binding}

This section provides the experimental details and supplementary results for
Binding Self-Distillation in Section~\ref{sec:binding}.

\subsection{Experimental Setup and Entity Selection}
\label{app:binding_setup}

Binding Self-Distillation is evaluated in a per-entity setting: a separate
lightweight adapter is trained and evaluated for each molecule. Parameters are
therefore not shared across molecules.

To ensure that the lexical route already exposes factual knowledge before
binding, we construct model-specific sets using therapeutic indication.
Starting from the multi-relation molecule collection, we retain a molecule
only when the model correctly retrieves its therapeutic indication through the
trivial name. Eligibility is determined only by indication; molecular target,
mechanism of action (MoA), and ATC class are not additionally filtered for
name-conditioned correctness.

\begin{table}[t]
\centering
\caption{\textbf{Model-specific Binding Self-Distillation sets.} The reported sets are the fixed entity sets selected by the original BSD
entity-selection pipeline using therapeutic-indication retrieval through the
trivial name.}

\label{tab:binding_entity_sets}
\small
\begin{tabular}{lr}
\toprule
Model & Entities \\
\midrule
ChemDFM-v2.0-14B         & 389 \\
Qwen2.5-14B-Instruct     & 390 \\
Llama-3.1-8B-Instruct    & 389 \\
ChemDFM-v1.5-8B          & 47 \\
Qwen2.5-7B-Instruct      & 351 \\
Meta-Llama-3-8B-Instruct & 378 \\
Qwen2-72B-Instruct       & 435 \\
\bottomrule
\end{tabular}
\end{table}

The reported Binding Self-Distillation experiments use the fixed entity sets
selected by the original experimental pipeline. In that pipeline, responses
were matched against accepted ChEMBL indication headings after lowercasing,
punctuation and possessive removal, whitespace normalization.

\subsection{Teacher--Student Construction}
\label{app:binding_prompts}

For a molecule with SMILES \(s\), trivial name \(n\), and generic continuation
stem \(u\), the student receives only the structural identifier,

\[
x_S =
\texttt{The molecule with SMILES }s\;u,
\]

whereas the teacher receives the same context augmented with the lexical
identity,

\[
x_T =
\texttt{The molecule with SMILES }s
\texttt{, which is known as ``}n\texttt{'', }u.
\]

Teacher and student are two forward passes through the same pretrained model.
The teacher remains frozen, while a LoRA adapter \(\phi\) on the student is
optimized using

\[
\mathcal{L}_{\mathrm{bind}}
=
D_{\mathrm{KL}}
\left(
p_{\theta}(\cdot\mid x_T)
\,\Vert\,
p_{\theta,\phi}(\cdot\mid x_S)
\right).
\]

The KL divergence is computed over the full vocabulary at the final prompt
position. No teacher continuation is generated and no factual answer is
provided as a training target. Binding Self-Distillation therefore transfers
the model's own predictive behavior induced by lexical identity rather than
training on external factual labels.

At every optimization step, six generic continuation stems are sampled
uniformly with replacement from a pool of 24. These include stems such as
\texttt{is a compound that}, \texttt{is known for},
\texttt{is a drug that}, \texttt{acts on}, and
\texttt{is associated with}. The exact downstream evaluation prompts are
excluded from the training pool. Some stems nevertheless cue broad biomedical
relation families, so we interpret the resulting gains as cross-prompt
transfer across multiple relations rather than strictly relation-held-out
generalization.

\paragraph{Anchor regularization.}
Each binding intervention additionally uses eight unrelated anchor molecules
sampled without replacement from the remaining eligible entities using
\texttt{random.Random(0)}. At each optimization step, an anchor is paired with
a newly sampled generic stem. The frozen model provides the original
next-token distribution for that prompt, and an additional KL term with
weight \(1.0\) regularizes the adapted model toward the original behavior.

\subsection{Optimization Configuration}
\label{app:binding_optimization}

For ChemDFM-v2.0-14B, ChemDFM-v1.5-8B,
Qwen2.5-7B-Instruct, Qwen2.5-14B-Instruct,
Llama-3.1-8B-Instruct, and Meta-Llama-3-8B-Instruct,
LoRA is applied to the \texttt{gate\_proj}, \texttt{up\_proj}, and
\texttt{down\_proj} modules in MLP layers 20--24, with target layer 24.
We use rank \(r=8\), scaling \(\alpha=16\), zero dropout, AdamW with zero
weight decay, learning rate \(10^{-3}\), and 120 optimization steps per
entity. Each step uses six binding contexts and eight anchor contexts.

\begin{table}[t]
\centering
\caption{\textbf{Binding Self-Distillation configuration for the six
verified local runs.}}
\label{tab:binding_hparams}
\small
\begin{tabular}{lc}
\toprule
Hyperparameter & Value \\
\midrule
Editing mode & One adapter per entity \\
Target layer & 24 \\
LoRA layers & 20--24 \\
LoRA modules & gate/up/down projection \\
Rank \(r\) & 8 \\
Scaling \(\alpha\) & 16 \\
Dropout & 0 \\
Optimizer & AdamW \\
Learning rate & \(1\times10^{-3}\) \\
Weight decay & 0 \\
Optimization steps & 120 \\
Binding contexts / step & 6 \\
Anchor molecules & 8 \\
Anchor weight & 1.0 \\
Objective & Full-vocabulary next-token KL \\
\bottomrule
\end{tabular}
\end{table}

Qwen2-72B-Instruct uses target layer 48 (relative depth \(0.60\)).

\subsection{Evaluation}
\label{app:binding_evaluation}

We evaluate four factual relations:

\begin{center}
\small
\begin{tabular}{ll}
\toprule
Relation & Continuation ending \\
\midrule
Indication & \texttt{is primarily used to treat} \\
Target & \texttt{has as its primary molecular target} \\
MoA & \texttt{acts by the following mechanism:} \\
ATC & \texttt{belongs to the therapeutic class} \\
\bottomrule
\end{tabular}
\end{center}

For each available relation, we evaluate:

\[
\begin{aligned}
\text{SMILES:}\quad&
\texttt{The molecule with SMILES \{s\} \{ending\}},\\
\text{Bridge:}\quad&
\texttt{The molecule with SMILES \{s\}, which is known as "\{name\}",
\{ending\}},\\
\text{Name:}\quad&
\texttt{\{Name.capitalize()\} \{ending\}}.
\end{aligned}
\]

Generation is greedy with at most 24 new tokens. Generated answers and accepted
targets are normalized by lowercasing, punctuation and possessive removal,
whitespace normalization, and light depluralization. A whole-phrase match to
any accepted target is counted as correct. For molecular target and MoA, we
additionally accept an answer covering at least \(60\%\) of the non-stopword
content tokens of an accepted target.

Let \(R_e\) denote the available factual relations for entity \(e\). We report

\[
F_c
=
\frac{1}{N}
\sum_{e=1}^{N}
\sum_{r\in R_e}
\mathbb{1}
\left[
\operatorname{hit}(e,r,c)
\right],
\]

where \(F_c\) is the mean number of correctly retrieved available relations
per entity under condition \(c\). Relations without an accepted answer are not
queried.

\subsection{Cross-Model Results}
\label{app:binding_all_models}

Table~\ref{tab:binding_crossmodel_full} reports relation-level results.

\begin{table*}[t]
\centering
\caption{Relation-wise Binding Self-Distillation results on the fixed entity sets selected by the
original BSD entity-selection pipeline. Values are retrieval accuracy (\%) through SMILES before
and after the per-entity binding intervention.}
\label{tab:binding_crossmodel_full}
\small
\setlength{\tabcolsep}{3.5pt}
\begin{tabular}{lrrrrrrrrr}
\toprule
& \multicolumn{2}{c}{Indication}
& \multicolumn{2}{c}{Target}
& \multicolumn{2}{c}{MoA}
& \multicolumn{2}{c}{ATC}
& \\
\cmidrule(lr){2-3}
\cmidrule(lr){4-5}
\cmidrule(lr){6-7}
\cmidrule(lr){8-9}
Model
& Before & After
& Before & After
& Before & After
& Before & After
& Facts/entity \\
\midrule
ChemDFM-v2.0-14B
& 2.1 & 51.4
& 0.0 & 27.6
& 0.0 & 5.0
& 3.3 & 11.4
& 0.05 $\rightarrow$ 0.87 \\

Qwen2.5-14B-Instruct
& 4.1 & 13.6
& 1.4 & 43.6
& 0.6 & 20.8
& 0.0 & 10.3
& 0.06 $\rightarrow$ 0.73 \\

Llama-3.1-8B-Instruct
& 0.3 & 38.8
& 0.7 & 19.9
& 1.2 & 8.4
& 1.1 & 7.9
& 0.03 $\rightarrow$ 0.68 \\

ChemDFM-v1.5-8B
& 25.5 & 42.6
& 5.1 & 23.1
& 17.1 & 12.2
& 6.8 & 13.6
& 0.51 $\rightarrow$ 0.85 \\

Qwen2.5-7B-Instruct
& 1.1 & 36.2
& 0.4 & 32.3
& 0.0 & 19.9
& 2.7 & 9.1
& 0.04 $\rightarrow$ 0.85 \\

Meta-Llama-3-8B-Instruct
& 0.3 & 42.3
& 0.7 & 21.8
& 0.6 & 13.0
& 0.3 & 8.8
& 0.02 $\rightarrow$ 0.78 \\
\bottomrule
\end{tabular}
\end{table*}

Across all six models, aggregate factual access increases after Binding
Self-Distillation. The gains extend beyond therapeutic indication to molecular
target, mechanism of action, and ATC class, supporting transfer beyond the
relation used to construct the eligible entity sets.

\subsection{Recovery Relative to Explicit Lexical Access}
\label{app:binding_bridge}

We compare the learned binding intervention with the inference-time lexical
bridge, which explicitly supplies the correct trivial name. We define bridge
recovery as

\[
100\,
\frac{
F_{\mathrm{post,SMILES}}-F_{\mathrm{pre,SMILES}}
}{
\max(
F_{\mathrm{pre,bridge}}-F_{\mathrm{pre,SMILES}},
10^{-9})
}.
\]

\begin{table}[t]
\centering
\caption{Binding Self-Distillation versus explicit lexical access on the fixed entity sets selected
by the original BSD entity-selection pipeline. Bridge gives factual access when the correct name is
explicitly supplied. Recovery measures the fraction of the SMILES-to-bridge gain recovered without
the name at inference time.}
\label{tab:binding_bridge_full}
\small
\setlength{\tabcolsep}{4pt}
\begin{tabular}{lrrrr}
\toprule
Model & \(n\) & Facts/entity & Bridge & Recovery \\
\midrule
ChemDFM-v2.0-14B
& 389 & 0.05 $\rightarrow$ 0.87 & 1.01 & 86\% \\
Qwen2-72B-Instruct
& 435 & 0.01 $\rightarrow$ 0.85 & 1.34 & 63\% \\
\bottomrule
\end{tabular}
\end{table}

For ChemDFM-v2.0-14B, Binding Self-Distillation recovers \(86\%\) of the
factual-access gain provided by explicitly supplying the lexical identity.
For Qwen2-72B-Instruct, it recovers \(63\%\). In both cases, inference after
training uses SMILES alone.

\subsection{Retention and Locality}
\label{app:binding_retention}

We additionally verify that the increased factual access need not come from
destroying behavior that already works. For ChemDFM-v2.0-14B, benzene outputs
are preserved on \(94.1\%\) of the retention probes. Name-conditioned
indication accuracy remains \(74.8\%\); target accuracy changes from
\(56.2\%\) to \(55.2\%\), ATC accuracy from \(16.0\%\) to \(16.3\%\), and
neighboring-molecule factual correctness from \(75.9\%\) to \(76.2\%\).

For Qwen2-72B-Instruct, exact-string comparison substantially understates
retention because the model frequently rephrases its answers. The underlying
benzene yes/no verdict is retained in \(92.9\%\) of cases, benzene correctness
against RDKit changes from \(83.9\%\) to \(86.9\%\), and neighboring-molecule
factual correctness remains \(80.5\%\).

\section{Generality Beyond Chemistry: Protein-Domain Replication}
\label{app:biology}

Our main experiments concern molecules, raising the possibility that
identifier-dependent knowledge access is specific to SMILES or structural
notation. We therefore repeat the key behavioral and causal tests in a
qualitatively different domain: proteins. A protein can be addressed through
a familiar lexical identifier such as its gene symbol or through an exact
    UniProt accession~\citep{ahmad2025uniprot} such as \texttt{P04637}. Unlike SMILES, an accession is an
opaque database key and encodes no biological structure. A similar asymmetry
therefore cannot be attributed to difficulty interpreting structural notation.

\subsection{Dataset Construction and Evaluation}
\label{app:biology_setup}

We query the UniProt REST API~\citep{ahmad2025uniprot} using
\texttt{(reviewed:true) AND (organism\_id:9606)}, requesting accession,
primary gene, recommended protein name, subcellular-location annotation, and
disease annotation. The data-generation script does not preserve an
authoritative UniProt release identifier or download timestamp.

For each returned entry, we retain the primary accession, the primary gene
symbol from the first gene record, and the recommended full protein name;
entries missing any of these fields are discarded. We additionally extract
the first recognized subcellular location and the first disease identifier.
An entry is retained if either annotation is available, so a recognized
location is not required for inclusion. No explicit accession, gene-symbol,
or protein-name deduplication is subsequently applied.

This produces 16,388 entries with 16,388 unique accessions and 16,336
distinct gene symbols. Of these, 15,925 have a recognized subcellular
location. ChemDFM-v2.0-14B is evaluated on all 16,388 entries. For the other
four models, entries are sorted by accession and the first 1,000 are used;
this is deterministic truncation rather than random sampling. The experiment
scripts do not preserve a reason for this 1,000-entry limit.

Subcellular location is mapped to the ordered label set
\[
\{
\text{nucleus},
\text{cytoplasm},
\text{cell membrane},
\text{mitochondrion},
\text{endoplasmic reticulum},
\text{golgi apparatus},
\text{secreted},
\text{lysosome},
\text{peroxisome},
\text{cytoskeleton}
\}.
\]
After normalization, the mapper returns the first label in this order whose
text occurs in a UniProt location annotation; the first recognized annotation
is retained. There is no additional alias table or multi-label evaluation.
The \texttt{cytoskeleton} class does not occur in the resulting data, leaving
nine observed labels. In the full set, the largest class is cytoplasm
(\(4{,}349/15{,}925=27.31\%\)); among the first 1,000 entries, 979 have
locations and the largest class is cell membrane
(\(281/979=28.70\%\)).

\paragraph{Prompts and scoring.}
All behavioral prompts have the form
\begin{quote}
\small\ttfamily
Consider \{phrase\}.\\
\{question\}\\
Answer in English. \{format\} Do not explain.
\end{quote}
The identifier-dependent phrase is one of
\begin{quote}
\small\ttfamily
the human protein encoded by the gene \{symbol\}\\
the human protein \{protein\_name\}\\
the human protein with UniProt accession \{accession\}\\
the human protein with UniProt accession \{accession\}, encoded by the gene
\{symbol\}\\
the human protein with UniProt accession \{accession\}, encoded by the gene
\{decoy\_symbol\}.
\end{quote}

For identification, the question is
\texttt{What is the official gene symbol for this protein?} with
\texttt{Reply with the gene symbol only.}
Thus, ``Symbol ID,'' ``Protein-name ID,'' and ``Accession ID'' all measure
recovery of the official gene symbol when conditioning on the corresponding
identifier.

For factual retrieval, the question is
\texttt{What is its primary subcellular location?}, and the model is asked to
reply with one of the ten location labels above. Bridge and conflicting-label
conditions use the gene symbol as the lexical identifier.

Generation is greedy with at most 32 new tokens and prompt truncation at
1,024 tokens. Models supporting chat templates additionally receive the
system message \texttt{You are a helpful assistant.}

The scorer rejects empty or refusal-like outputs and normalized responses
longer than 12 words. Identification succeeds when the gold gene symbol
appears as a case-insensitive whole phrase; no gene-symbol aliases are used.
For location, generated text and the stored gold label are lowercased,
possessives and punctuation are removed, whitespace is normalized, and light
depluralization is applied. A whole-phrase occurrence of the gold label is
then counted as correct. If several locations are mentioned, the response
therefore succeeds whenever the stored gold label appears.

\subsection{Identifier-Dependent Protein Knowledge}
\label{app:biology_behavior}

Table~\ref{tab:biology_access_full} reports the complete behavioral results.
We distinguish bridge recovery for \emph{identification} from bridge recovery
for \emph{subcellular location}; these are separate evaluations.

\begin{table*}[t]
\centering
\caption{\textbf{Identifier-dependent access also appears for proteins.}
Identification columns report recovery of the official gene symbol.
``Loc.\ via Acc.'' is subcellular-location accuracy through the accession.
ID and Loc.\ Bridge report the fraction of the respective
accession--gene-symbol gap recovered when the correct gene symbol is supplied
alongside the accession. Decoy ID measures whether identification follows an
incorrect gene symbol in a conflicting-identifier prompt. All values except
Entries are percentages.}
\label{tab:biology_access_full}
\small
\setlength{\tabcolsep}{3.2pt}
\resizebox{\textwidth}{!}{
\begin{tabular}{lrrrrrrrr}
\toprule
Model
& Entries
& Symbol ID
& Name ID
& Accession ID
& Loc.\ via Acc.
& ID Bridge
& Loc.\ Bridge
& Decoy ID \\
\midrule
ChemDFM-v2.0-14B
& 16,388 & 99.5 & 66.7 & 0.0 & 23.9 & 99.9 & 100.1 & 99.4 \\

ChemDFM-v1.5-8B
& 1,000 & 5.5 & 18.8 & 0.0 & 0.0 & 20.0 & 5.6 & 1.0 \\

Qwen2.5-14B-Instruct
& 1,000 & 98.0 & 60.8 & 0.0 & 22.0 & 100.7 & 101.0 & 98.6 \\

Qwen2.5-7B-Instruct
& 1,000 & 86.4 & 51.8 & 0.0 & 21.9 & 98.3 & 101.4 & 85.3 \\

Llama-3.1-8B-Instruct
& 1,000 & 90.8 & 60.3 & 0.0 & 9.3 & 98.9 & 97.6 & 90.9 \\
\bottomrule
\end{tabular}
}
\end{table*}

Accession-to-identity accuracy is effectively zero for every model. In
contrast, four models identify proteins reliably from gene symbols
(\(86.4\)--\(99.5\%\)); ChemDFM-v1.5-8B has substantially weaker lexical
protein knowledge. Location retrieval through the accession is likewise weak,
ranging from \(0.0\%\) to \(23.9\%\).

Supplying the correct gene symbol nearly eliminates this asymmetry for the four
models with strong lexical knowledge. For identification, bridge recovery is
\(98.3\)--\(100.7\%\); for subcellular location it is
\(97.6\)--\(101.4\%\). Values slightly above \(100\%\) occur when the
accession-plus-symbol prompt performs marginally better than the pure
gene-symbol prompt.

For a fact \(f\), bridge recovery is computed as
\[
\mathrm{BridgeRecovery}_f
=
100\,
\frac{
A_{\mathrm{acc+symbol},f}
-
A_{\mathrm{acc},f}
}{
\max(
A_{\mathrm{symbol},f}
-
A_{\mathrm{acc},f},
10^{-9}
)
}.
\]
For example, Qwen2.5-14B-Instruct identifies
\(0/1000\), \(987/1000\), and \(980/1000\) proteins from accession,
accession-plus-symbol, and symbol prompts, respectively, producing
\(100.7\%\) identification recovery.

We additionally pair each accession with an incorrect gene symbol. For every
entry, the decoy is sampled with seed 0 from the same evaluated population,
subject only to having a different gene-symbol string. For the four models
with strong lexical knowledge, identification follows this incorrect symbol
in \(85.3\)--\(99.4\%\) of cases. Specifically, the decoy is produced in
\(16{,}283/16{,}388\), \(986/1000\), \(853/1000\), and \(909/1000\)
cases for ChemDFM-v2.0-14B, Qwen2.5-14B-Instruct,
Qwen2.5-7B-Instruct, and Llama-3.1-8B-Instruct, respectively.
ChemDFM-v1.5-8B follows it in only \(10/1000\) cases, consistent with its
weak gene-symbol route.

Thus, the access asymmetry and conflicting-identifier behavior do not require
a structural representation such as SMILES: they also arise between a familiar
lexical identifier and an opaque exact database key.

\subsection{Activation Patching in the Protein Domain}
\label{app:biology_patching}

We next repeat the causal intervention from Section~\ref{sec:patching}.
Unlike the behavioral experiment, patching does not select examples according
to generated-answer correctness. It begins with accession-sorted entries having
a mapped location and uses the raw-completion prompts
\[
\begin{aligned}
x_{\mathrm{src}} &=
\texttt{The human protein encoded by the gene \{symbol\}
is primarily located in the},\\
x_{\mathrm{tgt}} &=
\texttt{The human protein with UniProt accession \{accession\}
is primarily located in the}.
\end{aligned}
\]
The gold completion is the canonical location label with a leading space.

For each item, we compute the teacher-forced mean log-probability per gold
token under the lexical source and accession target. We retain examples for
which the lexical route exceeds the accession route by at least \(0.20\) nats
per token. Examples are sorted by this gap, capped at 80 per gold label, and
then capped at 1,000 total examples. Because only nine location labels occur
and eligibility varies by model, the resulting sets contain
607--689 examples.

At layer \(L\), we capture the residual-stream output of decoder block
\texttt{layers[L]} at the \emph{last prompt token} of the lexical source and
overwrite the corresponding last-token state in the accession computation.
Thus, the intervention patches one residual-stream position after decoder
block \(L\); it does not patch the identifier token span itself. Layer
indexing is zero-based.

Let \(s_i\) and \(t_i\) denote source and untouched-target mean gold-token
log-probabilities for item \(i\), and \(p_{i,L}\) the corresponding score
after patching. Normalized recovery is
\[
R_{i,L}
=
\frac{
p_{i,L}-t_i
}{
\max(s_i-t_i,10^{-6})
}.
\]
Reported recovery is \(100\) times the arithmetic mean of \(R_{i,L}\) over
items, without clipping. Recovery can therefore exceed \(100\%\) when the
patched target assigns greater probability to the gold continuation than the
lexical source itself.

For the control, seed 0 selects another item from the same patching set with
a different gold location. Its source state is patched identically, giving
\[
C_{i,L}
=
\frac{
c_{i,L}-t_i
}{
\max(s_i-t_i,10^{-6})
}.
\]
The same control assignment is reused across layers. Negative control values
therefore indicate that an incompatible entity state reduces the gold
probability below that of the untouched accession route.

\begin{table*}[t]
\centering
\caption{\textbf{Activation patching in the protein domain.}
Peak is the layer maximizing mean normalized recovery after the complete
coarse-to-fine sweep; relative depth is peak layer divided by model depth.
Rank is evaluated at the coarse-sweep peak (see text). Rank top-1 reports
whether the correct location has the highest teacher-forced score among the
nine observed location classes.}
\label{tab:biology_patching_full}
\small
\setlength{\tabcolsep}{4pt}
\begin{tabular}{lrrrrrrr}
\toprule
Model
& Items
& Peak
& Rel. depth
& Recovery
& Control
& Rank layer
& Rank top-1 \\
\midrule
ChemDFM-v2.0-14B
& 689 & 47 & 0.98 & 89.0 & -104.2 & 40 & \(11.6\rightarrow50.7\) \\

ChemDFM-v1.5-8B
& 607 & 25 & 0.78 & 109.9 & -113.7 & 28 & \(12.7\rightarrow62.6\) \\

Qwen2.5-14B-Instruct
& 677 & 39 & 0.81 & 94.6 & -125.7 & 40 & \(11.8\rightarrow55.5\) \\

Qwen2.5-7B-Instruct
& 674 & 24 & 0.86 & 105.3 & -137.3 & 24 & \(11.9\rightarrow37.8\) \\

Llama-3.1-8B-Instruct
& 680 & 31 & 0.97 & 88.0 & -130.1 & 24 & \(11.8\rightarrow38.1\) \\
\bottomrule
\end{tabular}
\end{table*}

Patching the matched lexical state yields \(88.0\)--\(109.9\%\) normalized
recovery across all five models, whereas different-location controls strongly
decrease the gold score. The intervention therefore provides causal evidence
that information produced along the lexical route is sufficient to improve
prediction from the otherwise weak accession route.

\paragraph{Ranking analysis.}
As a complementary threshold-free measure, we score the complete gold strings
for every location represented in the patching set using mean
teacher-forced log-probability per token. Although the benchmark defines ten
possible labels, \texttt{cytoskeleton} is absent from the data, so the executed
ranking analysis contains nine candidates and uniform chance is \(11.1\%\).
The correct location initially ranks first only \(11.6\)--\(12.7\%\) of the
time, but rises to \(37.8\)--\(62.6\%\) after patching.

The implementation computes this ranking at the \emph{coarse-sweep peak},
rather than at the final dense-sweep maximum. We therefore report its actual
evaluation layer separately in Table~\ref{tab:biology_patching_full} and do
not treat the ranking values as measurements at the final peak layer.

\paragraph{Layer selection and localization.}
The patching sweep first evaluates layers
\(0,4,8,\ldots\) below each model's depth. It selects the coarse layer with
highest mean normalized recovery, then evaluates all previously untested
layers within \(\pm8\) layers of this coarse peak. The final reported peak is
the layer with maximum mean recovery over the complete saved sweep; rank and
control metrics do not influence selection.

The effective layers are substantially later than in chemistry. Chemistry
peaks at relative depths \(0.50\)--\(0.75\), whereas protein patching peaks
at \(0.78\)--\(0.98\). At these depths, especially \(0.97\)--\(0.98\) for
Llama-3.1-8B-Instruct and ChemDFM-v2.0-14B, the intervention occurs close to
the output of the network. We therefore interpret the protein experiment as
evidence of \emph{causal sufficiency}, not as a replication of the
chemistry-specific localization pattern.

\end{document}